\documentclass[10pt,twocolumn,letterpaper]{article}

\usepackage{cvpr}              %
\makeatletter
\@namedef{ver@everyshi.sty}{}
\makeatother
\usepackage{tikz}

\usepackage[square,numbers,sort&compress]{natbib}
\usepackage{enumitem}
\usepackage{algpseudocode}
\usepackage[font=small,labelfont=bf]{caption}
\usepackage{array}
\usepackage{multirow}
\usepackage{booktabs}
\usepackage{algorithm}
\usepackage{subcaption}
\usepackage[normalem]{ulem}
\usepackage{xparse}
\usepackage{pifont}
\usepackage{bm}
\usepackage{threeparttable}
\usepackage{etoolbox}

\usepackage{listings}
\usepackage{mwe} %
\usepackage{makecell}
\usepackage{color, colortbl}
\usepackage{tabularx}
\usepackage{pifont}%
\usepackage[accsupp]{axessibility}

\usepackage[scaled=0.85]{DejaVuSansMono}

\definecolor{citecolor}{HTML}{0071bc}
\usepackage[breaklinks=true,bookmarks=false,colorlinks,bookmarks=false, citecolor=citecolor, pagebackref]{hyperref}

\usepackage[capitalize]{cleveref}
\crefname{section}{\S}{\S\S}
\crefname{subsection}{\S}{\S\S}
\crefformat{table}{Table~#2#1#3}
\crefformat{figure}{Figure~#2#1#3}
\crefformat{equation}{Eq~#2#1#3}

\usepackage{graphicx}
\usepackage{amsmath}
\usepackage{amssymb}
\usepackage{pgfplots}
\pgfplotsset{compat=1.16}
\usepgfplotslibrary{groupplots}
\usetikzlibrary{patterns}

\newlength\savewidth\newcommand\shline{\noalign{\global\savewidth\arrayrulewidth
  \global\arrayrulewidth 1pt}\hline\noalign{\global\arrayrulewidth\savewidth}}

\newlength\thinwidth

\definecolor{Gray}{gray}{0.92}
\definecolor{DarkGray}{gray}{0.5}
\newcolumntype{x}{>{\columncolor{Gray}}c}
\newcolumntype{H}{>{\setbox0=\hbox\bgroup}c<{\egroup}@{}}
\definecolor{LightCyan}{rgb}{0.88,1,1}
\definecolor{altRowColor}{gray}{0.92}
\definecolor{highlightRowColor}{rgb}{0.9, 0.9, 1}
\newcommand{\colorrow}{\rowcolor{highlightRowColor}}

\newcommand{\highlightcell}{\cellcolor{highlightRowColor}}
\newcommand{\colorcell}{\cellcolor{Gray}}

\definecolor{GrayNumber}{gray}{0.5}
\definecolor{GrayXMark}{gray}{0.7}
\newcommand{\cmark}{\ding{51}}%
\newcommand{\xmark}{ {\color{GrayXMark} \ding{55}} } %

\definecolor{ImageDark}{rgb}{0,0.3,0.8}
\definecolor{VideoDark}{rgb}{.5,.0,.5}
\definecolor{DepthDark}{rgb}{0,.5,0}
\definecolor{AudioDark}{rgb}{0.11764705882352941, 0.5647058823529412, 1.0}
\definecolor{ThermalDark}{rgb}{0.8823529411,0.63725490196,0.0156862745}
\definecolor{IMUDark}{rgb}{0.6235294117647059, 0.27058823529411763, 0.4627450980392157}
\colorlet{Image}{ImageDark!20!white}
\colorlet{Video}{VideoDark!20!white}
\colorlet{Depth}{DepthDark!20!white}
\colorlet{Audio}{AudioDark!20!white}
\colorlet{ImageLight}{ImageDark!70!white}
\colorlet{VideoLight}{VideoDark!70!white}
\colorlet{DepthLight}{DepthDark!70!white}
\colorlet{AudioLight}{AudioDark!70!white}
\newcommand{\symbolHt}{1.5em}
\newcommand{\imageChar}{%
  \begingroup\normalfont
  \includegraphics[height=\symbolHt]{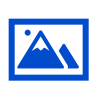}%
  \endgroup
}
\newcommand{\videoChar}{%
  \begingroup\normalfont
  \includegraphics[height=\symbolHt]{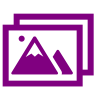}%
  \endgroup
}

\newcommand{\audioChar}{%
  \begingroup\normalfont
  \includegraphics[height=\symbolHt]{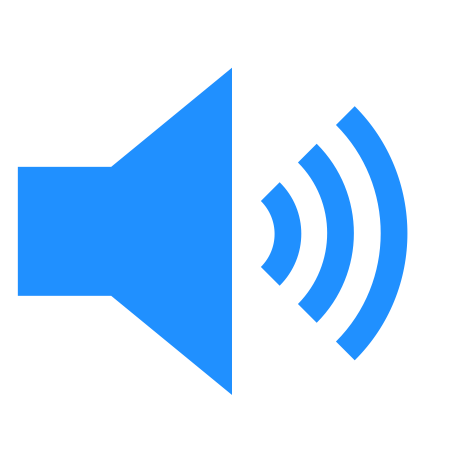}%
  \endgroup
}
\newcommand{\depthChar}{%
  \begingroup\normalfont
  \includegraphics[height=\symbolHt]{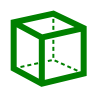}%
  \endgroup
}
\newcommand{\thermalChar}{%
  \begingroup\normalfont
  \includegraphics[height=\symbolHt]{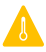}%
  \endgroup
}
\newcommand{\imuChar}{%
  \begingroup\normalfont
  \includegraphics[height=\symbolHt]{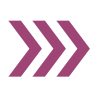}%
  \endgroup
}

\newcolumntype{i}{>{\columncolor{Image}}c}
\newcolumntype{v}{>{\columncolor{Video}}c}
\newcolumntype{d}{>{\columncolor{Depth}}c}
\newcolumntype{a}{>{\columncolor{Audio}}c}
\newcolumntype{I}{>{\columncolor{ImageLight}}c}
\newcolumntype{V}{>{\columncolor{VideoLight}}c}
\newcolumntype{D}{>{\columncolor{DepthLight}}c}
\newcolumntype{A}{>{\columncolor{AudioLight}}c}
\newcolumntype{E}{>{\columncolor{highlightRowColor}}c}
\newcommand{\tablestyle}[2]{\setlength{\tabcolsep}{#1}\renewcommand{\arraystretch}{#2}\centering\footnotesize}

\newcommand{\OURS}{\textsc{ImageBind}\xspace}
\newcommand{\Ours}{\OURS}

\newcommand{\bq}{\mathbf{q}}
\newcommand{\bk}{\mathbf{k}}

\newcommand{\bI}{\mathbf{I}}
\newcommand{\bM}{\mathbf{M}}
\newcommand{\setimage}{\mathcal{I}}

\newcommand{\setmodality}{\mathcal{M}}

\newcommand{\vit}{ViT\xspace}
\newcommand{\vitS}{ViT-S\xspace}
\newcommand{\vitB}{ViT-B\xspace}
\newcommand{\vitL}{ViT-L\xspace}
\newcommand{\vitH}{ViT-H\xspace}

\newcommand{\imnet}{ImageNet\xspace}
\newcommand{\imnetShort}{IN1K\xspace}

\newcommand{\placesThree}{Places-365\xspace}
\newcommand{\placesThreeShort}{P365\xspace}

\newcommand{\cocoShort}{COCO\xspace}

\newcommand{\sunrgbd}{SUN RGB-D\xspace}
\newcommand{\sunrgbdShort}{SUN\xspace}
\newcommand{\sunrgbdDepth}{SUN Depth-only\xspace}
\newcommand{\sunrgbdDepthShort}{SUN-D\xspace}

\newcommand{\nyuDepth}{NYU-v2 Depth-only\xspace}
\newcommand{\nyuDepthShort}{NYU-D\xspace}

\newcommand{\kineticsShort}{K400\xspace}

\newcommand{\msrvtt}{MSR-VTT 1k-A\xspace}
\newcommand{\msrvttShort}{MSR-VTT\xspace}

\newcommand{\ego}{Ego4D\xspace}
\newcommand{\egoShort}{Ego4D\xspace}

\newcommand{\audioset}{Audioset\xspace}
\newcommand{\audiosetShort}{AS\xspace}
\newcommand{\audiosetAudio}{Audioset Audio-only\xspace}
\newcommand{\audiosetAudioShort}{AS-A\xspace}
\newcommand{\audiocaps}{AudioCaps\xspace}
\newcommand{\audiocapsShort}{AudioCaps\xspace}
\newcommand{\esc}{ESC-50\xspace}
\newcommand{\escShort}{ESC\xspace}
\newcommand{\vggsound}{VGGSound\xspace}
\newcommand{\vggsoundShort}{VGGS\xspace}
\newcommand{\clotho}{Clotho\xspace}
\newcommand{\clothoShort}{Clotho\xspace}

\newcommand{\llvip}{LLVIP\xspace}
\newcommand{\llvipShort}{LLVIP\xspace}

\def\confName{CVPR}
\def\confYear{2023}

\begin{document}

\title{\Ours: One Embedding Space To Bind Them All}

\author{
  Rohit Girdhar$^{*}$ \quad \quad Alaaeldin El-Nouby$^{*}$ \quad \quad Zhuang Liu \quad \quad Mannat Singh  \\
  \quad \quad Kalyan Vasudev Alwala \quad \quad Armand Joulin \quad \quad Ishan Misra$^{*}$ \\
  FAIR, Meta AI \\
  {\small \url{https://facebookresearch.github.io/ImageBind}}
}

\twocolumn[{%
\renewcommand\twocolumn[1][]{#1}%
\maketitle
\begin{center}
    \centering
    \captionsetup{type=figure}
    \vspace{-0.2cm}
    \includegraphics[width=\linewidth]{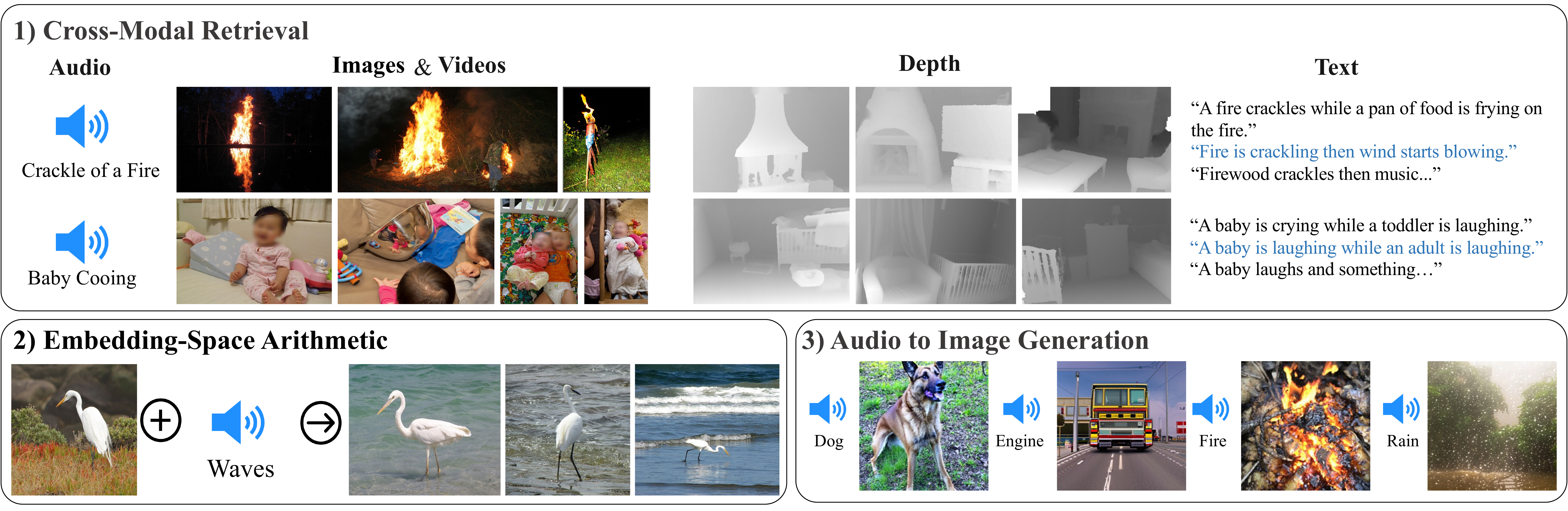}
    \captionof{figure}{
    \textbf{\OURS's joint embedding space enables novel multimodal capabilities.}
    By aligning six modalities' embedding into a common space, \OURS enables: \textbf{1)} Cross-Modal Retrieval, which shows \emph{emergent} alignment of modalities such as audio, depth or text, that aren't observed together. \textbf{2)} Adding embeddings from different modalities naturally composes their semantics.
    And \textbf{3)} Audio-to-Image generation, by using our audio embeddings with a pre-trained DALLE-2~\cite{ramesh2022hierarchical} decoder designed to work with CLIP text embeddings.
}
\label{fig:teaser}
\end{center}%
}]
\makeatletter{\renewcommand*{\@makefnmark}{}
\footnotetext{$^*$Equal technical contribution.}\makeatother}

\begin{abstract}
   \vspace{-0.2cm}
We present \OURS, an approach to learn a joint embedding across six different modalities - images, text, audio, depth, thermal, and IMU data.
We show that all combinations of paired data are not necessary to train such a joint embedding, and only image-paired data is sufficient to bind the modalities together.
\OURS can leverage recent large scale vision-language models, and extends their zero-shot capabilities to new modalities just by using their natural pairing with images.
It enables novel emergent applications `out-of-the-box' including cross-modal retrieval, composing modalities with arithmetic, cross-modal detection and generation.
The emergent capabilities improve with the strength of the image encoder and we set a new state-of-the-art on emergent zero-shot recognition tasks across modalities, outperforming specialist supervised models.
Finally, we show strong few-shot recognition results outperforming prior work, and that \OURS serves as a new way to evaluate vision models for visual and non-visual tasks.

\end{abstract}

\section{Introduction}
\label{sec:intro}

A single image can bind together many experiences -- an image of a beach can remind us of the sound of waves, the texture of the sand, a breeze, or even inspire a poem.
This `binding' property of images offers many sources of supervision to learn visual features, by aligning them with any of the sensory experiences associated with images.
Ideally, for a single joint embedding space, visual features should be learned by aligning to all of these sensors.
However, this requires acquiring all types and combinations of paired data with the same set of images, which is infeasible.

Recently, many methods learn %
image features %
aligned with text~\cite{schuhmann2021laion,schuhmann2022laion5b,radford2021learning,mahajan2018exploring,yu2022coca,yuan2021florence,jia2021scaling,alayrac2022flamingo}, audio~\cite{morgado2021audio,patrick2020multi,owens,bachmann2022multimae,tian2019contrastive,arandjelovic2017look} \etc.
These methods use a single pair of modalities or, at best, a few visual modalities. However, the final embeddings are limited to the pairs of modalities used for training.
Thus, video-audio embeddings cannot directly be used for image-text tasks and vice versa.
A major obstacle in learning a true joint embedding is the absence of large quantities of multimodal data where all modalities are present together.

In this paper, we present {\bf \OURS}, which learns a single shared representation space by leveraging multiple types of image-paired data.
It does not need datasets where all modalities co-occur with each other.
Instead, we leverage the binding property of images and we show that just aligning each modality's embedding to image embeddings leads to an emergent alignment across all of the modalities.
In practice, \OURS leverages web-scale (image, text) paired data and combines it with naturally occurring paired data such as (video, audio), (image, depth) \etc to learn a single joint embedding space.
This allows \OURS to implicitly align the text embeddings to other modalities such as audio, depth \etc, enabling zero-shot recognition capabilities on that modality without explicit semantic or textual pairing. %
Moreover, we show that it can be initialized with large-scale vision-language models such as CLIP~\cite{radford2021learning}, thereby leveraging the rich image and text representations of these models.
Thus, \OURS can be applied to a variety of different modalities and tasks with little training.

We use large-scale image-text paired data along with naturally paired `self-supervised' data across four new modalities - audio, depth, thermal, and Inertial Measurement Unit (IMU) readings -- and show strong emergent zero-shot classification and retrieval performance on tasks for each of these modalities.
These emergent properties improve as the underlying image representation is made stronger.
On audio classification and retrieval benchmarks, \OURS's emergent zero-shot classification matches or outperforms specialist models trained with direct audio-text supervision on benchmarks like \escShort, \clotho, \audiocaps.
\OURS representations also outperform specialist supervised models on few-shot evaluation benchmarks.
Finally, we show that \OURS's joint embeddings can be used for a wide variety of compositional tasks as illustrated in~\cref{fig:teaser}, including cross-modal retrieval, combining embeddings via arithmetic, detecting audio sources in images, and generating images given audio input.

\section{Related Work}
\label{sec:related}

\OURS builds upon several advances in vision-language, multimodal, and self-supervised research. \par \noindent
\textbf{Language Image Pre-training.}
Training images jointly with linguistic signals like words or sentences has been shown to be an effective method for
zero-shot, open-vocabulary recognition and text to image retrieval \cite{frome2013devise, faghri2017vse,
socher2014grounded, kiros2014unifying}. Language as supervision can further be used for learning strong video
representations \cite{alayrac2020self, miech2019howto100m, miech2020end}.
Joulin \etal~\cite{joulin2016learning} show that using large-scale image dataset with noisy captions yields strong
visual features.
Recently, CLIP~\cite{radford2021learning}, ALIGN~\cite{jia2021scaling} and Florence~\cite{yuan2021florence}
collect large collections of image and text pairs and train models to embed image and language inputs in a joint space
using contrastive learning, exhibiting impressive zero-shot performance. CoCa~\cite{yu2022coca} adds an image
captioning objective on top of the contrastive loss for improved performance. Flamingo~\cite{alayrac2022flamingo}
handles arbitrarily interleaved images and texts, and achieves state of the art on many few-shot learning benchmarks.
LiT~\cite{zhai2022lit} adopts contrastive training for fine-tuning and observes freezing image encoders works the best.
This prior line of works mostly considers image and text, while our work enables zero-shot recognition on multiple
modalities.

\par \noindent \textbf{Multi-Modal Learning.}
Our work binds multiple modality representations in a joint embedding space. Prior works explored joint training of
multiple modalities in a supervised \cite{girdhar2022omnivore,anonymous2021polyvit} or self-supervised contexts
\cite{tian2019contrastive, girdhar2022omnimae, wang2021bevt, morgado2021audio, arandjelovic2017look}.
 The success of image and language pre-training methods such as CLIP has inspired approaches that revisits learning
deep semantic representations through matching other modalities with linguistic inputs. Various methods adapt CLIP to
extract semantically strong video representations \cite{xue2022clip, luo2021clip4clip, fang2021clip2video,
lin2022frozen}. Most related to our method, Nagrani \etal~\cite{nagrani2022learning} create a weakly-labeled dataset
for paired video-audio and captions that allows for training multi-modal video-audio encoder to match textual features
resulting in strong audio and video retrieval and captioning performance. AudioCLIP~\cite{guzhov2021audioclip} adds
audio as an additional modality into a CLIP framework, enabling zero-shot audio classification.
In contrast, \OURS does not require explicit paired data between all modalities and instead leverages
image as a natural weak supervision for unifying modalities.

\par \noindent \textbf{Feature Alignment} Pre-trained CLIP models have been utilized as teachers to supervise other
models due to the strength of its visual representations \cite{wei2022contrastive, peng2022beit, liu2022exploring}.
Moreover, CLIP joint image and text embedding space has also been leveraged for a variety of zero-shot tasks like
detection~\cite{zhou2022detecting,gu2021open}, segmentation~\cite{li2022language}, mesh
animation~\cite{youwang2022clipactor} \etc showing the power of joint embedding spaces.
PointCLIP~\cite{zhang2022pointclip} finds a pre-trained CLIP encoder can be used for 3D recognition by projecting a
point cloud to a number of 2D depth map views, which in turn are encoded using CLIP visual encoder. In multilingual
neural machine translation, a similar phenomenon to the emergence behavior of \OURS is commonly observed and utilized:
if languages are trained in the same latent space through learned implicit bridging, translation can be done between
language pairs on which no paired data is provided~\cite{johnson2017google,lample2017unsupervised}.

\section{Method}
\label{sec:method}

\begin{figure*}[t]
    \centering
    \includegraphics[width=\linewidth]{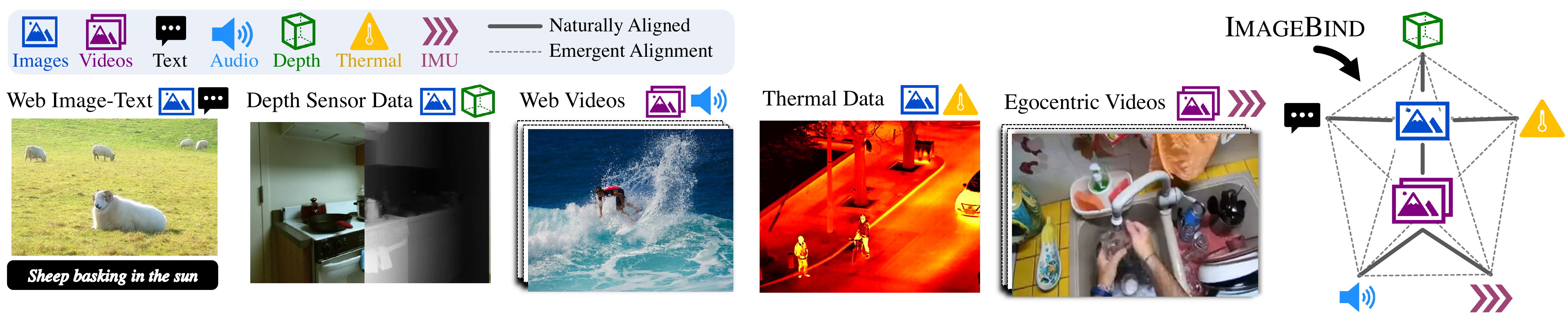}
    \caption{\textbf{\OURS overview.}
        Different modalities occur naturally aligned in different data sources, for instance images+text and video+audio in web data, depth or thermal information with images, IMU data in videos captured with egocentric cameras, \etc. \OURS links all these modalities in a common embedding space, enabling new emergent alignments and capabilities.
    }
    \label{fig:approach}
\end{figure*}

Our goal is to learn a single joint embedding space for all modalities by using images to bind them together. We align
each modality's embedding to image embeddings, such as text to image using web data and IMU to video using video data captured from egocentric cameras with IMU.
We show that the resulting embedding space has a powerful emergent zero-shot behavior that automatically
associates pairs of modalities without seeing any training data for that specific pair.
We illustrate our approach %
in~\cref{fig:approach}.

\subsection{Preliminaries}

\par \noindent \textbf{Aligning specific pairs of modalities.} Contrastive learning~\cite{hadsell2006dimensionality} is
a general technique for learning an embedding space by using pairs of related examples (positives) and unrelated
examples (negatives).
Using pairs of aligned observations, contrastive learning can \textbf{align pairs of modalities} such as (image,
text)~\cite{radford2021learning}, (audio, text)~\cite{guzhov2021audioclip}, (image, depth)~\cite{tian2019contrastive},
(video, audio)~\cite{morgado2021audio} \etc. However, in each case, the joint embeddings are trained and evaluated
using the same pairs of modalities. Thus, (video, audio) embeddings are not directly applicable for text-based tasks
while (image, text) embeddings cannot be applied for audio tasks.

\par \noindent \textbf{Zero-shot image classification using text prompts.} CLIP~\cite{radford2021learning} popularized
a `zero-shot' classification task based on an aligned (image, text) embedding space. This involves constructing a
list of text descriptions that describe the classes in a dataset.
An input image is classified based on its similarity to the text descriptions in the embedding space.
Unlocking such zero-shot classification for other modalities requires specifically training using paired text data,
\eg, (audio, text)~\cite{guzhov2021audioclip} or (point-clouds, text)~\cite{zhang2022pointclip}.
In contrast, \OURS unlocks zero-shot classification for modalities \emph{without} paired text data.

\subsection{Binding modalities with images}

\OURS uses pairs of modalities ($\setimage, \setmodality$), where $\setimage$ represents images and $\setmodality$ is
another modality, to learn a single joint embedding. We use large-scale web datasets with (image, text) pairings
that span a wide range of semantic concepts. Additionally, we use the natural, self-supervised pairing of other
modalities -- audio, depth, thermal, and Intertial Measurement Unit (IMU) -- with images.

Consider the pair of modalities ($\setimage, \setmodality$) with aligned observations.
Given an image $\bI_i$ and its corresponding observation in the other modality $\bM_i$, we encode them into normalized
embeddings: $\bq_i = f(\bI_i)$ and $\bk_i = g(\bM_i)$ where $f,g$ are deep networks. The embeddings and the encoders
are optimized using an InfoNCE~\cite{oord2018representation} loss:
\begin{equation}
    L_{\setimage,\setmodality} = - \log \frac{\exp(\bq_{i}^{\intercal} \bk_i/\tau)}{\exp(\bq_{i}^{\intercal} \bk_i/\tau) + \sum_{j \neq i}\exp(\bq_{i}^{\intercal} \bk_j/\tau)},
    \label{eq:contrastive_loss}
\end{equation}
where $\tau$ is a scalar temperature that controls the smoothness of the softmax distribution and $j$ denotes unrelated
observations, also called `negatives'. We follow~\cite{wu2018unsupervised} and consider every example $j\neq i$ in the
mini-batch to be a negative. The loss makes the embeddings $\bq_i$ and $\bk_i$ closer in the joint embedding
space, and thus aligns $\setimage$ and $\setmodality$.
In practice, we use a symmetric loss $L_{\setimage,\setmodality} + L_{\setmodality,\setimage}$.

\par \noindent \textbf{Emergent alignment of unseen pairs of modalities.} \OURS uses modalities paired with images,
\ie, pairs of the form ($\setimage,\setmodality$) to align each the embeddings from each modality $\setmodality$ to
those from images.
We observe an emergent behavior in the embedding space that aligns two pairs of modalities
$(\setmodality_1,\setmodality_2)$ even though we only train using the pairs $(\setimage,\setmodality_1)$ and
$(\setimage,\setmodality_2)$. This behavior allows us to perform a wide variety of zero-shot and cross-modal retrieval
tasks without training for them. We achieve state-of-the-art zero-shot text-audio classification results without
observing a single sample of paired (audio, text).

\subsection{Implementation Details}
\label{sec:implementation_details}

\OURS is conceptually simple and can be implemented in many different ways. We deliberately choose a vanilla
implementation that is flexible and allows for an effective study and easy adoption. In~\cref{sec:ablation}, we present
design decisions that are critical for good emergent `binding'.

\par \noindent \textbf{Encoding modalities.} We use a Transformer architecture~\cite{vaswani2017attention} for all the
modality encoders. We use the Vision Transformer (\vit)~\cite{dosovitskiy2020image} for images.
Following~\cite{girdhar2022omnimae}, we use the same encoder for images and videos. We temporally
inflate~\cite{carreira2017quo} the patch projection layer of the \vit and use $2$ frame video clips sampled from $2$ seconds.
We follow~\cite{gong21b_interspeech} for encoding audio and convert a $2$ second audio sampled at 16kHz into spectrograms
using $128$ mel-spectrogram bins. As the spectrogram is also a 2D signal like an image, we use a \vit
with a patch size of $16$ and stride $10$. We treat thermal images and depth images as one-channel images and also use a \vit to encode them. We follow~\cite{girdhar2022omnivore} to convert depth into disparity maps for scale invariance.
We extract the IMU signal consisting of accelerometer and gyroscope measurements across the $X$, $Y$, and $Z$ axes.
We use $5$ second clips resulting in 2K time step IMU readings which are projected using a 1D convolution with a kernel size of $8$.
The resulting sequence is encoded using a Transformer.
Finally, we follow the text encoder design from CLIP~\cite{radford2021learning}.

We use separate encoders for images, text, audio, thermal images, depth images, and IMU. We add a modality-specific
linear projection head on each encoder to obtain a fixed size $d$ dimensional embedding, that is normalized and used in the
InfoNCE loss from~\cref{eq:contrastive_loss}. In addition to ease of learning, this setup allows us to also initialize
a subset of the encoders using pretrained models, \eg, the image and text encoder using CLIP~\cite{radford2021learning} or OpenCLIP~\cite{ilharco2021openclip}.

\begin{table}
    \centering
    \resizebox{\linewidth}{!}{
    \setlength{\tabcolsep}{2pt}
    \begin{tabular}{l|cccc}
        \bf Dataset  & \bf Task & \bf \#cls & \bf Metric & \bf \#test \\
        \hline
        {\color{AudioDark}{\audiosetAudio}} ({\color{AudioDark}\audiosetAudioShort})~\cite{audioset} & Audio cls. & 527 & mAP & 19048 \\
        {\color{AudioDark}{\escShort} 5-folds} ({\color{AudioDark}\escShort})~\cite{esc} & Audio cls. & 50 & Acc & 400 \\
        {\color{AudioDark}{\clotho}} ({\color{AudioDark}\clothoShort})~\cite{clotho} & Retrieval & - & Recall & 1045 \\
        {\color{AudioDark}{\audiocaps}} ({\color{AudioDark}\audiocapsShort})~\cite{audiocaps} & Retrieval & - & Recall & 796 \\
        {\color{AudioDark}{\vggsound}} ({\color{AudioDark}\vggsoundShort})~\cite{vggsound} & Audio cls. & 309 & Acc & 14073 \\
        \arrayrulecolor{DarkGray}\hline
        {\color{DepthDark}{\sunrgbdDepth}} ({\color{DepthDark}\sunrgbdDepthShort})~\cite{song2015sun} & Scene cls. & 19 & Acc & 4660 \\
        {\color{DepthDark}{\nyuDepth}} ({\color{DepthDark}\nyuDepthShort})~\cite{silberman2012indoor} & Scene cls. & 10 & Acc & 653 \\
        \arrayrulecolor{DarkGray}\hline
        {\color{ThermalDark}{\llvip}} ({\color{ThermalDark}\llvipShort})~\cite{llvip} & Person cls. & 2 & Acc & 15809 \\
        \arrayrulecolor{DarkGray}\hline
        {\color{IMUDark}{\ego}} ({\color{IMUDark}\egoShort})~\cite{ego4d} & Scenario cls. & 108 & Acc & 68865 \\
    \end{tabular}
    }
    \caption{\textbf{Emergent zero-shot classification datasets} for {\color{AudioDark} audio}, {\color{DepthDark} depth}, {\color{ThermalDark} thermal}, and {\color{IMUDark} Inertial Measurement Unit (IMU)} modalities.
    We evaluate \OURS without training for any of these tasks and without training on paired text data for these modalities.
    For each dataset, we report the task (classification or retrieval), number of classes (\#cls), metric for evaluation (Accuracy or mean Average Precision), and the number of test samples (\#test).
    }
    \label{tab:transfer_datasets_zs}
\end{table}
\section{Experiments}
\label{sec:experiments}
\begin{table*}[!t]
    \setlength{\tabcolsep}{4pt}
    \centering
    \resizebox{\textwidth}{!}{
    \begin{tabular}{l | cc | cc | EE | EEE | E | E}
    & \multicolumn{2}{c}{\color{ImageDark} \imageChar{}} & \multicolumn{2}{c}{\color{VideoDark} \videoChar{}} & \multicolumn{2}{c}{\highlightcell \color{DepthDark} \depthChar{}} & \multicolumn{3}{c}{\highlightcell \color{AudioDark} \audioChar{}}
    & \multicolumn{1}{c}{\highlightcell \color{ThermalDark} \thermalChar{}} & \multicolumn{1}{c}{\highlightcell \color{IMUDark} \imuChar{}}
    \\
    & {\color{ImageDark} \imnetShort} & {\color{ImageDark} \placesThreeShort}
    & {\color{VideoDark} \kineticsShort} & {\color{VideoDark} \msrvttShort}
    & {\color{DepthDark} \nyuDepthShort} & {\color{DepthDark} \sunrgbdDepthShort}
    & {\color{AudioDark} \audiosetAudioShort} & {\color{AudioDark} \vggsoundShort} & {\color{AudioDark} \escShort}
    & {\color{ThermalDark} \llvip}
    & {\color{IMUDark} Ego4D}
    \\
    \shline
    Random & 0.1 & 0.27 & 0.25 & 0.1 &10.0 & 5.26 & 0.62 & 0.32 & 2.75 & 50.0 & 0.9 \\ %

    \OURS &
     77.7 %
     & 45.4 %
     & 50.0  %
     & 36.1  %
    & 54.0 & 35.1 & %
    17.6 & 27.8 %
    & 66.9 %
    & 63.4 %
    & 25.0 \\ %

     {Text Paired}
     & -
     & -
     & -
     & -
     & {41.9$^*$} %
     & {25.4$^*$} %
     & {28.4$^\dagger$~\cite{guzhov2021audioclip}} %
     & {-} %
     & {68.6$^\dagger$~\cite{guzhov2021audioclip}}  %

     & {-} %
     & {-} %
     \\

    \hline

    {\color{DarkGray} Absolute SOTA}
    & {\color{DarkGray} 91.0}~\cite{yu2022coca} %
    & {\color{DarkGray} 60.7}~\cite{singh2022revisiting} %
    & {\color{DarkGray}89.9}~\cite{yan2022multiview} %
    & {\color{DarkGray} 57.7~\cite{xue2022clip}} %
    & {\color{DarkGray} 76.7}~\cite{girdhar2022omnivore} %
    & {\color{DarkGray} 64.9}~\cite{girdhar2022omnivore} %
    & {\color{DarkGray} 49.6~\cite{koutini2022efficient} %
    }
    & {\color{DarkGray} 52.5~\cite{kazakos2021slow}}  %
    & {\color{DarkGray} 97.0~\cite{chen2022htsat}}

    & {\color{DarkGray} -} %
    & {\color{DarkGray} -} %
    \\
    \end{tabular}}
    \caption{\textbf{Emergent zero-shot classification} of \OURS using text prompts \colorbox{highlightRowColor!150}{highlighted in blue}.
    \OURS aligns images with text, depth, audio, thermal and IMU modalities.
    The resulting embedding space can associate text embeddings with the non-image modalities, and leads to strong emergent zero-shot classification.
    We show strong performance even on non-visual modalities such as audio and IMU.
    We compare to `Text Paired' baselines wherever possible, which trains with paired text data for that modality. $^*$We use the OpenCLIP \vitH~\cite{ilharco2021openclip} on depth rendered as grayscale images.
    $^\dagger$\cite{guzhov2021audioclip} that uses \audiosetShort class names as supervision during training, and hence is not ``zero-shot''.
    Overall, \OURS shows strong emergent zero-shot performance, even compared to such upper bounds.
    We also report the absolute state-of-the-art (SOTA)  on each dataset for reference, which typically uses additional supervision, model ensembles \etc.
    We report the top-1 classification accuracy for all datasets except \msrvttShort (Recall@1) and \audiosetAudio (mAP).
    }
    \label{tab:emergent_zero_shot}
\end{table*}

We first describe the main experimental setup and provide full details in the supplement.

\par \noindent \textbf{Naturally paired modalities and datasets.} We use \OURS on six modalities - image/video, text, audio, depth, thermal images, and IMU. As described in~\cref{sec:implementation_details}, we treat videos as
2 frame images and process them the same as images. For the naturally available paired data, we use the (video,
audio) pairs from the \audioset dataset~\cite{audioset}, (image, depth) pairs from the \sunrgbd
dataset~\cite{song2015sun}, (image, thermal) pairs from the \llvip dataset~\cite{llvip} and (video, IMU) pairs from the
\ego dataset~\cite{ego4d}. For these pairs of modalities, we do not use any extra supervision like class
labels, text \etc. Since \sunrgbd and \llvip are relatively small, we
follow~\cite{girdhar2022omnivore} and replicate them 50$\times$ for training.

\par \noindent \textbf{Large scale image-text pairs.} We leverage image-text supervision from large-scale web
data~\cite{radford2021learning}. For ease of experimentation, we use pretrained models that are trained on billions
of (image, text) pairs. Specifically, we use the pretrained vision (\vitH 630M params) and text encoders (302M params) from
OpenCLIP~\cite{ilharco2021openclip,cherti2022scalinglaws}. %

\par \noindent \textbf{Encoders for each modality.} We convert audio into 2D mel-spectrograms~\cite{gong21b_interspeech}, and thermal and depth
modalities into 1 channel images and use \vitB, \vitS encoders respectively. The image and text
encoders are kept frozen during the \OURS training and the audio, depth, thermal,
and IMU encoders are updated.

\par \noindent \textbf{Emergent zero-shot \vs zero-shot.} Methods such as CLIP~\cite{radford2021learning},
AudioCLIP~\cite{guzhov2021audioclip} \etc train with modality pairs, (image, text) and (audio, text), to demonstrate
zero-shot classification using text-prompts for the same modality. In contrast, \OURS binds modalities together using
only image-paired data. Thus, just by training on (image, text) and (image, audio), \OURS can perform zero-shot
classification of audio using text prompts. As we do not directly train for this ability, we term it \emph{emergent}
zero-shot classification to distinguish it from methods that specifically train using paired
text-supervision for all modalities.

\par \noindent \textbf{Evaluation on downstream tasks.} We comprehensively evaluate \OURS on a many different
downstream tasks using different protocols.
We summarize the main datasets used for evaluation in~\cref{tab:transfer_datasets_zs}.

\subsection{Emergent zero-shot classification}
We evaluate \OURS on emergent zero-shot classification and use the text prompt templates from~\cite{radford2021learning} (full
details in~\cref{appendix:eval_details}). We report the results in~\cref{tab:emergent_zero_shot}. Each task
measures \OURS's ability to associate text embeddings to the other modalities without observing them together during
training.
Given the novelty of our problem setting, there are no ``fair'' baselines to compare
\OURS with. Nevertheless, we compare to prior work that uses text paired with certain modalities (\eg audio~\cite{nagrani2022learning,guzhov2021audioclip}), and for certain
``visual-like'' modalities such as depth and thermal, we use the CLIP model directly.
We also report the best reported supervised upper bound per benchmark.

\OURS achieves a high emergent zero-shot classification performance. On each benchmark, \OURS achieves
strong gains and even compares favorably to supervised specialist models trained for the specific modality and task.
These results demonstrate that \OURS aligns the modalities and implicitly transfers the text supervision associated
with images to other modalities like audio. In particular, \OURS shows strong alignment for non-visual modalities like
audio and IMU suggesting that their naturally available pairing with images is a powerful source of supervision. For
completeness, we also report the standard zero-shot image (ImageNet~\cite{ILSVRC15} - \imnetShort, \placesThree~\cite{Places205} - \placesThreeShort) and video (Kinetics400~\cite{kay2017kinetics} - \kineticsShort, \msrvtt~\cite{xu2016msr} - \msrvttShort) tasks. As the image \& text
encoders are initialized (and frozen) using OpenCLIP, these results match those of OpenCLIP.

\begin{table}[!t]
    \setlength{\tabcolsep}{3pt}
    \centering
    \resizebox{0.45\textwidth}{!}{
    \begin{tabular}{l| c | cc | cc | c}
    & Emergent & \multicolumn{2}{c}{\color{AudioDark} \clothoShort} &  \multicolumn{2}{c}{\color{AudioDark} \audiocapsShort} &  {\color{AudioDark} \escShort}
    \\
    && R@1 &  R@10  & R@1 & R@10 & Top-1
    \\
    \shline

    \multicolumn{3}{l}{\emph{Uses audio and text supervision}} \\
    AudioCLIP~\cite{guzhov2021audioclip} & \xmark  & \_ & \_ & \_ & \_ & {\bf 68.6} \\ %
    \midrule
    \multicolumn{3}{l}{\emph{Uses audio and text loss}} \\
    AVFIC~\cite{nagrani2022learning} & \xmark &
    3.0 & 17.5 &  %
    8.7 &  37.7  %
    & \_ \\
    \midrule
    \multicolumn{3}{l}{\emph{No audio and text supervision}} \\
    \OURS & \cmark &
    \textbf{6.0} & \textbf{28.4} & \textbf{9.3} & \textbf{42.3} %
    & 66.9 %
    \\ %
    \midrule
    \multicolumn{3}{l}{{\color{DarkGray}\emph{Supervised}}} \\
    {\color{DarkGray} AVFIC finetuned~\cite{nagrani2022learning}} & {\color{DarkGray} \xmark} & {\color{DarkGray}8.4} & {\color{DarkGray}38.6} & \_ & \_ & \_ \\ %
    {\color{DarkGray}ARNLQ~\cite{oncescu2021audio}} & {\color{DarkGray} \xmark} & {\color{DarkGray}12.6} & {\color{DarkGray}45.4} & {\color{DarkGray} 24.3} & {\color{DarkGray}72.1} & \_ %
    \end{tabular}}
    \caption{\textbf{Emergent zero-shot audio retrieval and classification.}
    We compare \OURS to prior work on zero-shot audio retrieval and audio classification.
    Without using audio-specific supervision, \OURS outperforms prior methods on zero-shot retrieval and has comparable performance on the classification task.
    \OURS's emergent zero-shot performance approaches those of specialist supervised models.
    }
    \label{tab:text_retrieval_audio}
\end{table}

\subsection{Comparison to prior work}
We now compare \OURS against prior work in zero-shot retrieval and classification tasks.

\par \noindent \textbf{Zero-shot text to audio retrieval and classification.} Unlike \OURS, prior work trains using
paired data for that modality, \eg, AudioCLIP~\cite{guzhov2021audioclip} uses (audio, text) supervision and
AVFIC~\cite{nagrani2021attention} uses automatically mined (audio, text) pairs. We compare their zero-shot text to
audio retrieval and classification performance to \OURS's emergent retrieval and classification
in~\cref{tab:text_retrieval_audio}.

\OURS significantly outperforms prior work on the audio text retrieval benchmarks. On the \clotho dataset, \OURS has
double the performance of AVFIC despite not using any text pairing for audio during training. Compared to the supervised
AudioCLIP model, \OURS achieves comparable audio classification performance on \escShort. Note that AudioCLIP uses class names from AudioSet as text targets for audio-text training, hence is referred to as `supervised'. \OURS's strong performance on all
three benchmarks validates its ability to align the audio and text modalities using images as a bridge.

\par \noindent \textbf{Text to audio and video retrieval.} We use the \msrvtt benchmark to evaluate the text to
audio and video retrieval performance in~\cref{tab:text_retrieval_vtt}. Only using audio, \OURS achieves strong
emergent retrieval performance compared to the video retrieval performance of prior work like MIL-NCE.
The text to video performance for our model is strong (36.1\% R@1 in~\cref{tab:emergent_zero_shot}) as it uses
OpenCLIP's vision and text encoders and outperforms many prior methods. However, combining the audio and video
modalities further boosts performance showing the utility of \OURS's features over an already strong retrieval model.

\begin{table}[!t]
    \setlength{\tabcolsep}{3pt}
    \centering
    \resizebox{0.45\textwidth}{!}{
    \begin{tabular}{l |c| c| ccc}
    & Modality & Emergent & \multicolumn{3}{c}{\color{VideoDark} \msrvttShort}
    \\
    &&& R@1 & R@5 & R@10
    \\
    \shline
    MIL-NCE~\cite{miech2021howto100m} & {\color{VideoDark}V} & \xmark & 8.6 & 16.9 & 25.8 \\ %
    SupportSet~\cite{patrick2020support} & {\color{VideoDark}V} & \xmark & 10.4 & 22.2 & 30.0 \\ %
    FIT~\cite{bain2021frozen} & {\color{VideoDark}V} & \xmark & 15.4 & 33.6 & 44.1 \\ %
    AVFIC~\cite{nagrani2022learning} & {\color{AudioDark}A}+{\color{VideoDark}V} & \xmark &
    19.4 & 39.5 & 50.3 \\ %
    \midrule
    
    \colorrow \OURS & {\color{AudioDark}A} & \cmark & 6.8 & 18.5 & 27.2 \\  %
    \OURS & {\color{AudioDark}A}+{\color{VideoDark}V} & \xmark & 36.8 & 61.8 & 70.0 \\ %
    
    \end{tabular}}
    \caption{\textbf{Zero-shot text based retrieval} on \msrvttShort 1K-A.
    We compare \OURS's emergent retrieval performance using audio and observe that it performs favorably to methods that use the stronger video modality for retrieval.
    }
    \label{tab:text_retrieval_vtt}
\end{table}

\subsection{Few-shot classification}
\begin{figure}[t!]
    \resizebox{0.49\linewidth}{!}{
        \begin{tikzpicture}
    \begin{axis}[
        xtick={0, 1, 2, 4, 8}, %
        legend pos=south east,
        xticklabels={0, 1, 2, 4, 8}, %
        xmin=0,
        grid=both,
        grid style={line width=.1pt, draw=gray!10},
        major grid style={line width=.2pt,draw=gray!50},
        minor tick num=2,
        axis x line*=bottom,
        axis y line*=left,
        width=\linewidth,
        ylabel style= {align=center, font=\large},
        xlabel style = {font=\large},
        ylabel={\escShort (Fold-1) Top-1},
        xlabel={\# shots per class},
        yticklabel style = {font=\large},
        xticklabel style = {font=\large},
        legend style={cells={align=left}, font=\large, fill=none, draw=none},
    ]
    \addplot[mark=o, very thick, AudioDark, mark options={solid}, line width=2.5pt, mark size=3.4pt] plot coordinates {
        (1, 62.0) %
        (2, 68.8) %
        (4, 81.3) %
        (8,  85.5) %
    };\label{pgf:audio_model_scaling_audiocaps:img}
    \addlegendentry{\OURS}

    \addplot[mark=square, very thick, gray!60, mark options={solid}, line width=2.5pt, mark size=3.4pt %
            ] plot coordinates {
        (1, 16.8) %
        (2, 19.5) %
        (4, 43.0) %
        (8,  56.3) %
    };\label{pgf:audio_model_scaling_audiocaps:img}
    \addlegendentry{AudioMAE~\cite{audiomae}}

    \addplot[mark=triangle, very thick, dashed, gray!60, mark options={solid}, line width=2.5pt, mark size=3.4pt  %
            ] plot coordinates {
        (1, 46.3) %
        (2, 62.5) %
        (4, 81.5) %
        (8,  87.5) %
    };\label{pgf:audio_model_scaling_audiocaps:img}
    \addlegendentry{Supervised~\cite{audiomae}}
    \addplot[mark=star, very thick, AudioDark, mark options={mark size=6pt, solid, line width=3pt}] plot coordinates {
        (0, 65.1) %
    };\label{pgf:audio_model_scaling_audiocaps:img}

    \end{axis}
\end{tikzpicture}
    } \hfill
    \resizebox{0.49\linewidth}{!}{
        \begin{tikzpicture}
    \begin{axis}[
        xtick={0, 1, 2, 4, 8}, %
        legend pos=south east,
        xticklabels={0, 1, 2, 4, 8}, %
        xmin=0,
        grid=both,
        grid style={line width=.1pt, draw=gray!10},
        major grid style={line width=.2pt,draw=gray!50},
        minor tick num=2,
        axis x line*=bottom,
        axis y line*=left,
        width=\linewidth,
        ylabel style= {align=center, font=\large},
        xlabel style = {font=\large},
        ylabel={\sunrgbdDepthShort Top-1},
        xlabel={\# shots per class},
        yticklabel style = {font=\large},
        xticklabel style = {font=\large},
        legend style={cells={align=left}, font=\large, fill=none, draw=none},
    ]
    \addplot[mark=o, very thick, DepthDark, mark options={solid}, line width=2.5pt, mark size=3.4pt] plot coordinates {
        (1, 21.3) %
        (2, 33.6) %
        (4, 35.6) %
        (8, 37.2) %
    };\label{pgf:depth_model_scaling_depthcaps:img}
    \addlegendentry{\OURS}

    \addplot[mark=square, very thick, gray!60, mark options={solid}, line width=2.5pt, mark size=3.4pt %
    ] plot coordinates {
        (1, 8.4) %
        (2, 13.4) %
        (4, 16.2) %
        (8, 19.9) %
    };\label{pgf:depth_model_scaling_depthcaps:img}
    \addlegendentry{MultiMAE~\cite{bachmann2022multimae}}

    \addplot[mark=star, very thick, DepthDark, mark options={mark size=6pt, solid, line width=3pt}] plot coordinates {
        (0, 35.1) %
    };\label{pgf:depth_model_scaling_depthcaps:img}

    \end{axis}
\end{tikzpicture}
    }
    \caption{\textbf{Few-shot classification on audio and depth.}
    We report the emergent zero-shot classification performance on each benchmark (denoted by $\star$).
    We train linear classifiers on fixed features for the $\ge\!1$-shot case.
    \textbf{(Left)} In all settings, \OURS outperforms the self-supervised AudioMAE model.
    \OURS even outperforms a supervised AudioMAE model upto 4 shot learning showing its strong generalization.
    \textbf{(Right)}
    We compare with the MultiMAE model trained with images, depth, and semantic segmentation masks.
    \OURS outperforms MultiMAE across all few-shot settings on few-shot depth classification.
    }
    \label{fig:low_shot}
\end{figure}
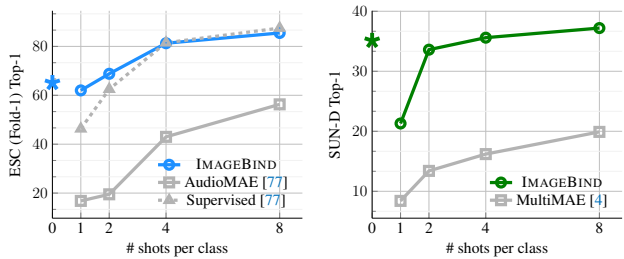
We now evaluate the label-efficiency of \OURS by evaluating on few-shot classification. We use the audio and depth
encoders from \OURS and evaluate them on audio and depth classification respectively in~\cref{fig:low_shot}. For
$\ge$1-shot results, we follow~\cite{radford2021learning,morgado2021audio} and train linear classifiers on fixed
features (details in~\cref{appendix:eval_details}).

On few-shot audio classification (\cref{fig:low_shot} left), we compare with (1)
self-supervised AudioMAE model trained on audio from Audioset and (2) a supervised AudioMAE model finetuned on audio
classification. Both baselines use the same capacity \vitB audio encoder
as \OURS. \OURS significantly outperforms the AudioMAE model on all settings with
gains of $\sim$40\% accuracy in top-1 accuracy on $\le$4-shot classification. \OURS also matches or outperforms the
supervised model on $\ge$1-shot classification. \OURS's emergent zero-shot performance surpasses the
supervised $\le$2-shot performance.

For few-shot depth classification, we compare with the multimodal MultiMAE~\cite{bachmann2022multimae} \vitB/16 model
trained on images, depth, and semantic segmentation data.
\OURS significantly outperforms MultiMAE across all the few-shot settings.
Altogether, these results show the strong generalization of \OURS audio and depth features trained with image alignment.

\subsection{Analysis and Applications}
\begin{figure}[t]
 \centering
    \captionsetup{type=figure}
    \includegraphics[width=\linewidth]{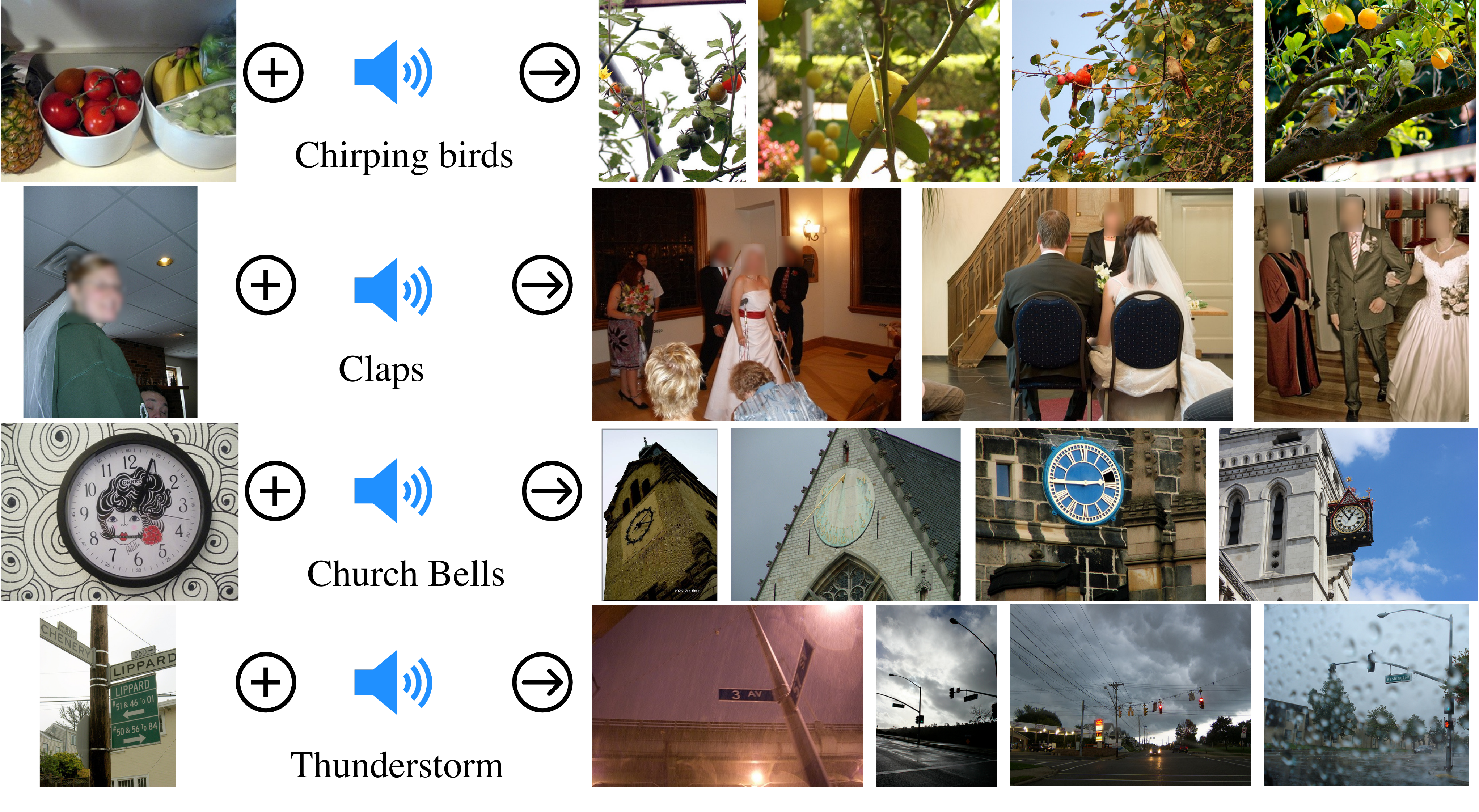}
    \captionof{figure}{\textbf{Embedding space arithmetic} where we add image and audio embeddings, and use them for image retrieval.
    The composed embeddings naturally capture semantics from different modalities.
    Embeddings from an image of fruits + the sound of birds retrieves images of birds surrounded by fruits.
    }
    \label{fig:image_audio_compose}
\end{figure}

\par \noindent \textbf{Multimodal embedding space arithmetic.}
We study whether \OURS's embeddings can be used to compose information across modalities.
In~\cref{fig:image_audio_compose}, we show image retrievals obtained by adding together image and audio embeddings.
The joint embedding space allows for us to compose two embeddings: \eg, image of fruits on a table + sound of chirping birds and retrieve an image that contains both these concepts, \ie, fruits on trees with birds.
Such \emph{emergent compositionality} whereby semantic content from different modalities can be composed will likely enable a rich variety of compositional tasks.

Without re-training, we can `upgrade' existing vision models that use CLIP embeddings to use \OURS embeddings from other modalities such as audio.
\par
\begin{figure}
    \includegraphics[width=\linewidth]{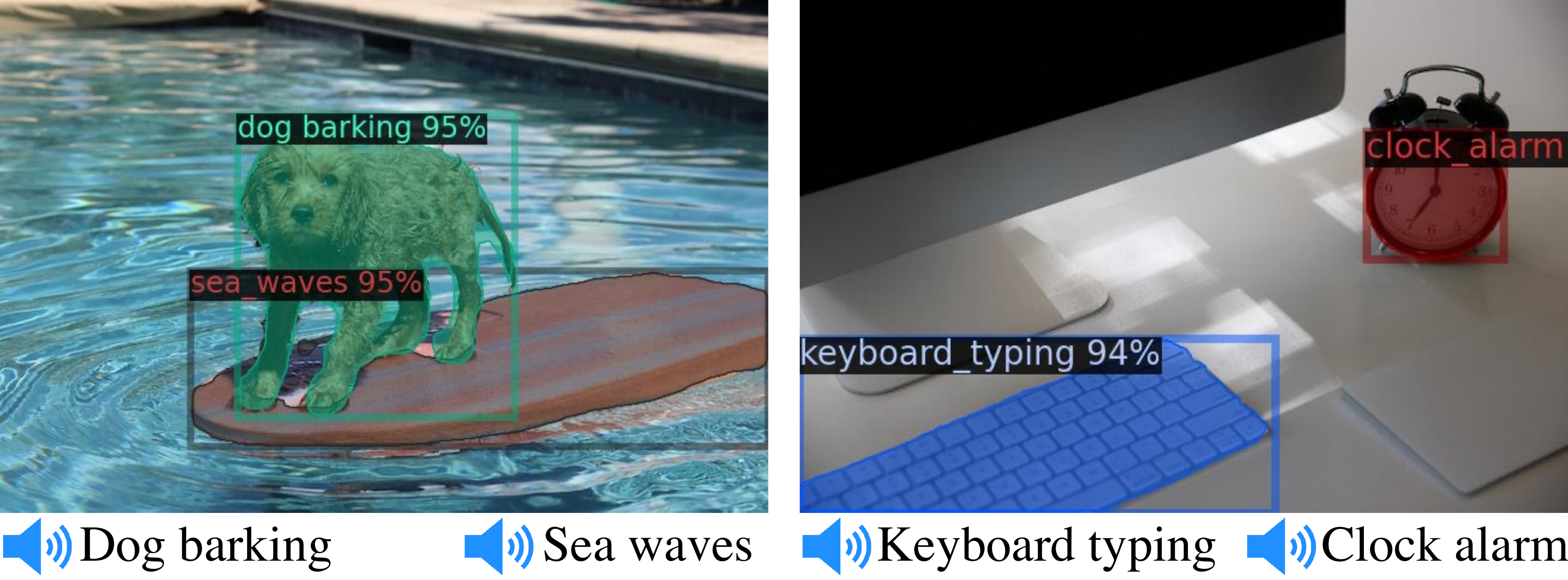}
    \caption{\textbf{Object detection with audio queries.}
    Simply replacing Detic~\cite{zhou2022detecting}'s CLIP-based `class' embeddings with our audio embeddings leads to an object detector promptable with audio.
    This requires no re-training of any model.
    }
    \label{fig:audio_detic}
\end{figure}

 \noindent \textbf{Upgrading text-based detectors to audio-based.}
We use a pretrained text-based detection model, Detic~\cite{zhou2022detecting}, and simply replace its CLIP-based `class' (text) embeddings with \OURS's audio embeddings.
Without training, this creates an `audio'-based detector that can detect and segment objects based on audio prompts.
As shown in~\cref{fig:audio_detic}, we can prompt the detector with the barking sound of a dog to localize a dog.

\par \noindent \textbf{Upgrading text-based diffusion models to audio-based.}
We use a pretrained DALLE-2~\cite{ramesh2022hierarchical} diffusion model (private reimplementation) and replace its prompt embeddings by our audio embeddings.
In~\cref{fig:teaser}, we observe that we can repurpose the diffusion model to generate plausible images using different types of sounds.

\maketitle

\begin{figure}[t!]
    \begin{subfigure}[b]{0.48\linewidth}
            \begin{tikzpicture}
    \begin{axis}[
        xtick={86, 307, 632},
        xticklabels={ViT-B, ViT-L, ViT-H},
        grid=both,
        grid style={line width=.1pt, draw=gray!10},
        major grid style={line width=.2pt,draw=gray!50},
        minor tick num=2,
        axis x line*=bottom,
        axis y line*=left,
        height=1.4in,
        width=\linewidth,
        ylabel style= {align=center},
        ylabel={\nyuDepthShort~\ref{pgf:depth_model_scaling_nyu:img}},
        yticklabel style = {font=\small},
        xticklabel style = {font=\footnotesize},
        legend style={cells={align=left}, font=\small},
    ]
    \addplot[mark=o, very thick, DepthDark] plot coordinates {
        (632, 50.8) %
        (307, 45.4) %
        (86,  43.6) %
    };\label{pgf:depth_model_scaling_nyu:img}
    \end{axis}
\end{tikzpicture}
    \end{subfigure}
    \hfill
    \begin{subfigure}[b]{0.48\linewidth}
            \begin{tikzpicture}
    \begin{axis}[
        xtick={86, 307, 632},
        xticklabels={ViT-B, ViT-L, ViT-H},
        grid=both,
        grid style={line width=.1pt, draw=gray!10},
        major grid style={line width=.2pt,draw=gray!50},
        minor tick num=2,
        axis x line*=bottom,
        axis y line*=left,
        height=1.4in,
        width=\linewidth,
        ylabel style= {align=center},
        ylabel={\escShort Fold-1~\ref{pgf:audio_model_scaling_esc:img}},
        yticklabel style = {font=\small},
        xticklabel style = {font=\small},
        legend style={cells={align=left}, font=\small},
    ]
    \addplot[mark=o, very thick, AudioDark] plot coordinates {
        (632, 65.1) %
        (307, 62.5) %
        (86,  61.3) %
    };\label{pgf:audio_model_scaling_esc:img}
    \end{axis}
\end{tikzpicture}
    \end{subfigure}
    \hfill
    \begin{subfigure}[b]{0.48\linewidth}
            \begin{tikzpicture}
    \begin{axis}[
        xtick={86, 307, 632},
        xticklabels={ViT-B, ViT-L, ViT-H},
        grid=both,
        grid style={line width=.1pt, draw=gray!10},
        major grid style={line width=.2pt,draw=gray!50},
        minor tick num=2,
        axis x line*=bottom,
        axis y line*=left,
        height=1.4in,
        width=\linewidth,
        ylabel style= {align=center},
        ylabel={\llvip~\ref{pgf:thermal_model_scaling_llvip:img}},
        yticklabel style = {font=\small},
        xticklabel style = {font=\footnotesize},
        legend style={cells={align=left}, font=\small},
    ]
    \addplot[mark=o, very thick, ThermalDark] plot coordinates {
        (632, 62.7) %
        (307, 60.5) %
        (86,  58.4) %
    };\label{pgf:thermal_model_scaling_llvip:img}
    \end{axis}
\end{tikzpicture}
    \end{subfigure}
    \begin{subfigure}[b]{0.48\linewidth}
            \begin{tikzpicture}
    \begin{axis}[
        xtick={86, 307, 632},
        xticklabels={ViT-B, ViT-L, ViT-H},
        grid=both,
        grid style={line width=.1pt, draw=gray!10},
        major grid style={line width=.2pt,draw=gray!50},
        minor tick num=2,
        axis x line*=bottom,
        axis y line*=left,
        height=1.4in,
        width=\linewidth,
        ylabel style= {align=center},
        ylabel={\ego~\ref{pgf:imu_model_scaling_ego4d:img}},
        yticklabel style = {font=\small},
        xticklabel style = {font=\footnotesize},
        legend style={cells={align=left}, font=\small},
    ]
    \addplot[mark=o, very thick, VideoDark] plot coordinates {
        (632, 25.0) %
        (307, 23.5) %
        (86,  17.3) %
    };\label{pgf:imu_model_scaling_ego4d:img}
    \end{axis}
\end{tikzpicture}
    \end{subfigure}
    \caption{\textbf{Scaling the image encoder} size while keeping the other modality encoders' size fixed.
    We measure the performance on the emergent zero-shot classification of depth, audio, thermal, and IMU modalities.
    Scaling the image encoder significantly improves the zero-shot classification results suggesting that a stronger visual representation improves the `binding' of modalities.
    \label{fig:vision_model_strength}
    }
\end{figure}
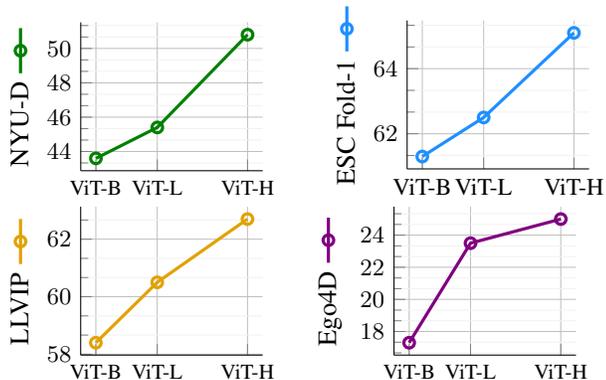
\section{Ablation Study}
\label{sec:ablation}

We investigate various design choices for learning a joint embedding space for different modalities. Since the ablation
experimental setup is similar to~\cref{sec:experiments}, we only note the main differences (full details
in~\cref{appendix:pretraining_details}). We report results on the \escShort fold-1 for the ablation study. We use a \vitB
encoder for the image, audio, depth, and thermal modalities by default and train them for 16 epochs (\vs 32 epochs in~\cref{sec:experiments}).
For IMU we use a lightweight 6 layer encoder with 512 dimensional width and 8 heads, and train it for 8 epochs.
The text encoder follows~\cite{radford2021learning} and is a twelve layer Transformer with a width of 512 dimensions.
We initialize the image and text encoder from the CLIP model~\cite{radford2021learning}.

\begin{table*}[!t]
	\centering
	\subfloat[
	\textbf{Temperature for loss}.
	\label{tab:ablate_contrastive_temp}
	]{
		\centering
		\begin{minipage}{0.25\linewidth}{\begin{center}
					\tablestyle{1.5pt}{1.2}
					\begin{tabular}{c|ccccc}
                        Temp $\rightarrow$ & Learn & 0.05 & 0.07 & 0.2 & 1.0 \\ [.1em]
                        \shline
                        {\color{DepthDark} \sunrgbdDepthShort} &
                         24.1 & 27.0 & 27.3 & \colorcell 26.7 & 28.0 \\
                         {\color{AudioDark} \escShort} &
                         54.8  &%
                         \colorcell 56.7  & %
                         52.4 & %
                         45.4 & %
                         24.3 \\ %
                    \end{tabular}
		\end{center}}\end{minipage}
	}
	\subfloat[
	\textbf{Projection Head}.
	\label{tab:ablate_proj_head}
	]{
		\begin{minipage}{0.25\linewidth}{\begin{center}
					\tablestyle{1.5pt}{1.2}
					\begin{tabular}{c|cc}
                        Proj head $\rightarrow$& Linear & MLP \\ [.1em]
                        \shline
                        {\color{DepthDark} \sunrgbdDepthShort} & \colorcell 26.7 %
                        & 
                        26.5 %
                        \\
                        {\color{AudioDark} \escShort} & \colorcell 56.7 &  %
                        51.0 %
                        \\
                    \end{tabular}
		\end{center}}\end{minipage}
	}
    \subfloat[
	\textbf{Training epochs}.
	\label{tab:ablate_epochs}
	]{
		\begin{minipage}{0.25\linewidth}{\begin{center}
            \tablestyle{1.5pt}{1.2}
            \begin{tabular}{c|ccc}
                Epochs $\rightarrow$ & 16 & 32 & 64 \\ [.1em]
                \shline
                {\color{DepthDark}\sunrgbdDepthShort} & 26.7 & 27.9 & \colorcell 29.9 \\
                {\color{AudioDark}\escShort} & 56.7 & 61.3 &
                 \colorcell 62.9 \\ %
            \end{tabular}
		\end{center}}\end{minipage}
	}
	\subfloat[
	\textbf{Data aug for image}.
	\label{tab:ablate_data_aug_image}
	]{
		\begin{minipage}{0.25\linewidth}{\begin{center}
					\tablestyle{1.5pt}{1.2}
					\begin{tabular}{c|ccc}
                        Data aug $\rightarrow$& Basic & Strong \\ [.1em]
                        \shline
                        {\color{DepthDark} \sunrgbdDepthShort} & 25.4 %
                        & \colorcell 26.7 %
                        \\
                        {\color{AudioDark} \escShort} & \colorcell 56.7  & %
                        22.6 \\ %
                    \end{tabular}
		\end{center}}\end{minipage}
	}
    
    \subfloat[
	\textbf{Spatial alignment of depth}.
	\label{tab:ablate_depth_alignment}
	]{
		\begin{minipage}{0.25\linewidth}{\begin{center}
					\tablestyle{1.5pt}{1.2}
					\begin{tabular}{c|cccc}
                        Spatial align $\rightarrow$& None & Aligned \\ [.1em]
                        \shline
                        {\color{DepthDark} \sunrgbdDepthShort} & 16.0 %
                        & \colorcell 26.7 \\  %
                    \end{tabular}
		\end{center}}\end{minipage}
	}
    \subfloat[
	\textbf{Depth data aug}.
	\label{tab:ablate_depth_data_aug}
	]{
		\begin{minipage}{0.25\linewidth}{\begin{center}
					\tablestyle{1.5pt}{1.2}
					\begin{tabular}{c|cccc}
                        Data aug $\rightarrow$& None & RandErase \\ [.1em]
                        \shline
                        {\color{DepthDark} \sunrgbdDepthShort} &  24.2 %
                        & \colorcell 26.7 \\  %
                    \end{tabular}
		\end{center}}\end{minipage}
	}
    \subfloat[
	\textbf{Temporal alignment of audio}.
	\label{tab:ablate_audio_alignment}
	]{
		\begin{minipage}{0.25\linewidth}{\begin{center}
					\tablestyle{1.5pt}{1.2}
					\begin{tabular}{c|cccc}
                        Temporal align$\rightarrow$& None & Aligned \\ [.1em]
                        \shline
                        {\color{AudioDark} \escShort} & 55.7 & %
                        \colorcell 56.7 \\  %
                    \end{tabular}
		\end{center}}\end{minipage}
	}
    \subfloat[
	\textbf{Audio data aug}.
	\label{tab:ablate_audio_data_aug}
	]{
		\begin{minipage}{0.25\linewidth}{\begin{center}
					\tablestyle{1.2pt}{1.2}
					\begin{tabular}{c|cccc}
                        Data aug $\rightarrow$& Basic & +Freq mask \\ [.1em]
                        \shline
                        {\color{AudioDark} \escShort} & 56.5 %
                        &
                        \colorcell 56.7 \\  %
                    \end{tabular}
		\end{center}}\end{minipage}
	}
    \caption{\textbf{Training loss and architecture} design decisions and their impact on emergent zero-shot classification.
    Settings for results in~\cref{sec:experiments} highlighted in \colorbox{Gray}{gray}.
    \textbf{(a)} A fixed temperature in the contrastive loss outperforms a learnable one for all modalities.
    \textbf{(b)} A linear projection head for computing the depth or audio embedding works better than an MLP head.
    \textbf{(c)} Longer training improves the zero-shot classification performance for both modalities.
    \textbf{(d)} Stronger image augmentation improves depth classification while basic augmentation significantly improves audio classification.
    \textbf{(e, f)} Using spatially aligned image and depth crops when training \OURS significantly improves performance.
    Similarly, RandErase augmentation is critical to good zero-shot classification on depth.
    \textbf{(g, h)} Temporally aligned audio and video matching gives improved performance and using frequency augmentation for audio gives a slight improvement.
    }
	\label{tab:ablate_hyperparameters}
 
\end{table*}

\subsection{Scaling the Image Encoder}
The central idea in \OURS is aligning the embeddings of all modalities to image embeddings. Thus, the image embeddings
plays a central role in the emergent alignment of unseen modalities and we study their effect on the emergent zero-shot
performance.
We vary the size of the image encoder and train an encoder for the depth, audio \etc modalities to match the image
representation. To isolate the effect of the image representation, we fix the size of the other modality encoders.
We use the pretrained CLIP (\vitB and \vitL) and OpenCLIP (\vitH) image and text encoders for this
experiment. Our results in~\cref{fig:vision_model_strength} show that \OURS's emergent zero-shot performance on all
modalities improves with better visual features.
For depth and audio classification, the stronger \vitH \vs the \vitB image encoder, provides a gain of 7\% and 4\%
respectively.
Thus, stronger visual features can improve recognition performance even on non-visual modalities.

\subsection{Training Loss and Architecture}
\label{sec:ablate_contrastive_arch}
We study the effect of the training design choices on the emergent zero-shot classification. We focus on two modalities
with different characteristics - depth which is visual and spatial, and audio which is non-visual and has a temporal
component. We found that studying these diverse modalities led to robust and transferable design decisions.

\par \noindent \textbf{Contrastive loss temperature.} We study the effect of the temperature $\tau$
(~\cref{eq:contrastive_loss}) in~\cref{tab:ablate_contrastive_temp}. We experiment with a learnable temperature
initialized to $0.07$ (parametrized in the log-scale) following~\cite{radford2021learning} \vs various values of fixed
temperatures. Unlike \cite{radford2021learning}, %
we observe that a fixed
temperature is best for depth, audio and IMU classification. Additionally, we see that a higher temperature is better
for training the depth, thermal, and IMU encoders, whereas a lower temperature works best for the audio modality.

\par \noindent \textbf{Projection head.} We vary the projection head used for each encoder from a linear layer to an
MLP with $768$ hidden dimensions. The results in~\cref{tab:ablate_proj_head} show that a linear projection performs
better for both modalities. This is in contrast to standard self-supervised methods like SimCLR~\cite{chen2020simple}
whose performance improves with MLP projection heads.

\par \noindent \textbf{Training epochs.} We vary the number training epochs and report the classification performance
in~\cref{tab:ablate_epochs}. Longer training consistently improves the emergent zero-shot performance for both
modalities across all datasets.

\par \noindent \textbf{Data augmentation for paired images.} During \OURS training, we augment images either using
basic augmentation (cropping, color jitter) or strong augmentation that additionally applies
RandAugment~\cite{cubuk2020randaugment} and RandErase~\cite{zhong2020random}. We specify the augmentation parameters
in~\cref{appendix:pretraining_details}. Stronger augmentation helps depth classification when training on the small
number of (image, depth) pairs from the \sunrgbd dataset. However, for audio, strongly augmenting the
video makes the task too challenging, leading to a significant drop  of 34\% on \escShort.

\par \noindent \textbf{Depth specific design choices.} We vary the type of spatial crops used for training
in~\cref{tab:ablate_depth_alignment}. Following CMC~\cite{tian2019contrastive}, we use two unaligned random crops from
the corresponding image and depth pair \vs our default choice of using spatially aligned random crops. Contrary to CMC,
we observe that random cropping severely degrades performance: more than 10\% on \sunrgbdDepthShort.
Unlike vanilla self-supervised learning, our image representations learned from image-text pairs are more semantic
and thus spatially misaligned crops hurt performance. In~\cref{tab:ablate_depth_data_aug}, we observe that
RandomErase~\cite{zhong2020random} boosts performance on depth classification.

\par \noindent \textbf{Audio specific design choices.} We train for video-audio alignment using temporally aligned
samples or unaligned samples and measure the final performance in~\cref{tab:ablate_audio_alignment}. Similar to the
depth classification observation, temporally aligned samples lead to better performance.
~\cref{tab:ablate_audio_data_aug} shows that using frequency masking augmentation for audio also provides a small boost
in performance.

\begin{table}[!t]
    \centering
    \setlength{\tabcolsep}{3pt}
    \renewcommand{\arraystretch}{1.05}
    \resizebox{\linewidth}{!}{
        \begin{tabular}{c| c c| c c}
              & \multicolumn{2}{c}{ \color{AudioDark} Audio Encoder (ESC)} & \multicolumn{2}{c}{\color{DepthDark} Depth Encoder (SUN)} \\
             Image Encoder & \color{AudioDark}  ViT-S & \color{AudioDark}  ViT-B & \color{DepthDark} ViT-S & \color{DepthDark} ViT-B \\
             \shline
             ViT-B  &
             52.8 & %
             56.7 &
             30.7 & %
             26.7 \\ %
             ViT-H  &
             54.8 & %
             \textbf{60.3} & %
             \textbf{33.3} & %
             29.5 \\ %
        \end{tabular}
    }
    \caption{\textbf{Capacity of the audio and depth encoders} and their impact on performance.
    A stronger image encoder improves performance for both audio and depth tasks.
    As the number of (image, depth) pairs is small, a smaller encoder improves performance for depth.
    For audio classification, a larger encoder is better.
    }
    \label{tab:capacity_matrix}
\end{table}
\par \noindent \textbf{Capacity of the audio and depth encoders} and their impact of the classification performance is
reported in~\cref{tab:capacity_matrix}. A smaller encoder for depth improves performance presumably because of the
relatively small size of the (image, depth) dataset. Conversely, we observe that larger audio encoder improves the
performance, particularly when paired with a high capacity image encoder.

\begin{table}[t]
    \centering
    \begin{tabular}{c|cccc}
        Batch size $\rightarrow$& 512 & 1k & 2k & 4k \\ [.1em]
        \shline
        {\color{DepthDark} \nyuDepthShort} &  \colorcell 47.3 & 46.5 & 43.0 & 39.9 \\ %
        {\color{AudioDark} \escShort} & 39.4 & 53.9 & \colorcell 56.7 & 53.9 \\ %
    \end{tabular}
    \caption{{\bf Effect of scaling batch size.} We found the optimal batch size for contrastive loss varied by the modality. For image-depth task, a smaller batch size was better, likely due to the small size and limited diversity of the original dataset. For audio-video task where we have a lot more positive and negative audio-video pairs, using a large batch size lead to better results.}\label{tab:batch_size}
\end{table}

{\bf \noindent Effect of batch size.} In~\cref{tab:batch_size} we evaluate the effect of batch size on the representation learned. As shown, the batch size can vary across modalities depending on the size and complexity of the corresponding pretraining datasets.

\par \noindent \textbf{\OURS to evaluate pretrained vision models} in~\cref{tab:different_image_encoders}.
We initialize the vision encoder using a pretrained model and keep it fixed.
We use image-paired data to align and train text, audio, and depth encoders (full details in~\cref{appendix:eval_details}).
Compared to the supervised DeiT model, the self-supervised DINO model is better at emergent zero-shot classification on both depth and audio modalities.
Moreover, the emergent zero-shot performance is not correlated with the pure vision performance on \imnet suggesting that these tasks measure different properties.
\OURS can serve as a valuable tool to measure vision models' strength on multimodal applications.

\begin{table}[t!]
    \centering
    \setlength{\tabcolsep}{3pt}
    \renewcommand{\arraystretch}{1.05}
    \resizebox{0.85\linewidth}{!}{
        \begin{tabular}{c| c | H c c | c c}
             &  {\color{ImageDark} \imnetShort}& \color{AudioDark}  \audiosetAudioShort & \color{AudioDark}  \vggsoundShort & \color{AudioDark}  \escShort & \color{DepthDark} \sunrgbdDepthShort & \color{DepthDark} \nyuDepthShort\\
             \shline
             DINO~\cite{caron2021emerging}  & 64.4 %
             & 3.1  %
             & 17.2  %
             & 44.7  %
             & 26.8 %
             & 48.8 %
             \\
             DeiT~\cite{touvron2021training}  & 74.4$^\dagger$ %
             & 1.4  %
             & 9.6 %
             & 25.0 %
             & 25.2 %
             & 48.0 %
        \end{tabular}
    }
    \caption{\textbf{\OURS as an evaluation tool.} We initialize (and fix) the image encoder with different methods and align other modalities.
    \OURS measures the impact of visual features on multimodal tasks.
    $^\dagger$ trained with \imnetShort supervision.
    }
    \label{tab:different_image_encoders}
\end{table}

\section{Discussion and Limitations}
\label{sec:discussion}

\OURS is a simple and practical way to train a joint embedding space using only image alignment. %
Our method leads to emergent alignment across all modalities which can be measured using cross-modal retrieval and text-based zero-shot tasks.
We enable a rich set of compositional multimodal tasks across different modalities, show a way to evaluate pretrained vision models for non-vision tasks and `upgrade' models like Detic and DALLE-2 to use using audio.
There are multiple ways to further improve \OURS.
Our image alignment loss can be enriched by using other alignment data, for instance other modalities paired with text, or with each other (\eg audio with IMU).
Our embeddings are trained without a specific downstream task, and thus lag the performance of specialist models.
More research into adapting general purpose embeddings for each task, including structured prediction tasks such as detection will be beneficial.
Finally, new benchmarks, \eg our emergent zero-shot task to measure emergent abilities of multimodal models, would help create exciting new applications.
Our model is a research prototype and cannot be readily used for real world applications (~\cref{appendix:rai}).

{\noindent \bf Acknowledgements:} Authors would like to thank Uriel Singer, Adam Polyak and Naman Goyal for their help with the DALLE-2 experiments, and the entire Meta AI team for many helpful discussions.

{\small
\bibliographystyle{ieee_fullname}
\bibliography{refs}
}
\clearpage
\appendix
\section{Datasets and Metrics}
\label{appendix:data_details}

{\bf \noindent \color{AudioDark} \audioset (\audiosetShort)}~\cite{audioset}.
This dataset is used for both training and evaluation. It contains 10s videos from YouTube annotated into 527 classes. It consists of 3 pre-defined splits, the balanced split with about 20K videos, test split with 18K videos, and an unbalanced training split with about 2M vidoes. For {\bf training}, we use the 2M unbalanced set without any labels, and only use it for audio-video matching.
For {\bf zero-shot evaluation} in~\cref{tab:emergent_zero_shot}, we use the test set and compute logits for each class using the textual class names along with the templates as described later in~\cref{appendix:zero_shot_eval}. The metric used is top-1 accuracy.

{\bf \noindent \color{AudioDark} \esc (\escShort)}~\cite{esc}.
We use this dataset for evaluating the learned representations in a zero-shot manner. The task here is ``Environmental Sound Classification'' (ESC). It consists of 2000 5s audio clips classified into 50 classes. It has pre-defined 5 fold evaluation, each consisting of 400 test audio clips. In this work, we compute 0-shot predictions on the evaluation set for each fold and report the 5-fold average performance. For ablations we use only the first fold for computational ease. The metric used is top-1 accuracy.

{\bf \noindent \color{AudioDark} \clotho (\clotho)}~\cite{clotho}.
This is a dataset of audio from the Freesound platform with textual descriptions. It consists of a dev and test set of 2893 and 1045 audio clips respectively, with each clip associated with 5 descriptions. We consider the text$\rightarrow$audio retrieval task, and consider each of the 5 associated captions as a separate test query and retrieve from the set of audio clips. The metric used is recall@$K$, where a given test query is assumed to be correctly solved if the ground truth audio is retrieved within the top-$K$ retrieved audio clips.

{\bf \noindent \color{AudioDark} \audiocaps (\audiocapsShort)}~\cite{audiocaps}.
This is a dataset of audiovisual clips from YouTube accompanied by textual descriptions. It consists of clips from the \audioset dataset as described earlier.
We use the splits as provided in~\cite{oncescu2021audio},\footnote{\url{https://www.robots.ox.ac.uk/~vgg/research/audio-retrieval/resources/benchmark-files/AudioCaps_retrieval_dataset.tar.gz}} which removes clips that overlap with the \vggsound dataset.
We end up with 48198 training, 418 validation and 796 test clips. We only use the test set for zero-shot evaluation of our model. The task is text$\rightarrow$audio retrieval, and evaluation is performed using recall@$K$.

{\bf \noindent \color{AudioDark} \vggsound (\vggsoundShort)}~\cite{vggsound}.
This dataset contains about 200K video clips of 10s length, annotated with 309 sound classes consisting of human actions, sound-emitting objects and human-object interactions.
We only use the audio from the test set (with 14073 clips) for 0-shot classification. The evaluation is done using the top-1 accuracy metric.

{\bf \noindent \color{DepthDark} \sunrgbd (\sunrgbdShort).}
We use the registered RGB and Depth maps provided in the \sunrgbd~\cite{song2015sun} dataset \texttt{train} set ($\sim$5K pairs) for training our model.
We follow~\cite{girdhar2022omnivore} to post process the depth maps in two steps - 1) we use in-filled depth values and 2) convert them to disparity for scale normalization.
This dataset is only used in training, so we do not use any metadata or class labels.

{\bf \noindent \color{DepthDark} \sunrgbdDepth (\sunrgbdDepthShort).}
We use only the $\sim$5K depth maps from the \texttt{val} split of the \sunrgbd~\cite{song2015sun} dataset and denote them as \sunrgbdDepth.
This dataset is only used for evaluation and we do not use the RGB images.
We process the depth maps similar to \sunrgbd (in-filled depth, converted to disparity).
We use the 19 scene classes in the dataset and use their class names for constructing the zero-shot classification templates.

{\bf \noindent \color{DepthDark} \nyuDepth (\nyuDepthShort).}
We use the 794 \texttt{val} set depth maps from the \nyuDepth~\cite{silberman2012indoor} dataset for evaluation only.
We post-process the depth similar to \sunrgbdDepth.
We use the 10 scene class names in the dataset.
The 10th scene class, called `other', correspond to 18 different semantic classes --
\texttt{['basement', 'cafe', 'computer lab', 'conference room', 'dinette', 'exercise room', 'foyer', 'furniture store', 'home storage', 'indoor balcony', 'laundry room', 'office kitchen', 'playroom', 'printer room', 'reception room', 'student lounge', 'study', 'study room']}.
For zero-shot evaluation, we compute the cosine similarity of the 10th class as the maximum cosine similarity among these 18 classnames.

{\bf \noindent \color{ThermalDark} \llvip (\llvipShort).}
The LLVIP dataset~\cite{llvip} consists of RGB image and Thermal (infrared low-light) image pairs.
The dataset was collected in an outdoor setting using fixed cameras observing street scenes and contains RGB images taken in a low-light paired with infrared images (8$\sim$14um frequency).
The RGB thermal pairs are registered in the dataset release.
For training, we use the \texttt{train} set with 12025 RGB image and thermal pairs.
For evaluation, we use the \texttt{val} set with 3463 pairs of RGB and thermal images.
Since the original dataset is designed for detection, we post process it for a binary classification task.
We crop out pedestrian bounding boxes and random bounding boxes (same aspect ratio and size as pedestrian) to create a balanced set of 15809 total boxes (7931 `person' boxes).
For zero-shot classification, we use the following class names for the `person' class - \texttt{['person', 'man', 'woman', 'people']}, and \texttt{['street', 'road', 'car', 'light', 'tree']} for the background class.

{\bf \noindent \color{IMUDark} \ego (\egoShort)}~\cite{ego4d}. For the \egoShort dataset, we consider the task of scenario
classification. There are 108 unique scenarios present in the 9,645 videos of the \egoShort dataset. We filter out all
videos annotated with more than one scenario which yields 7,485 videos with a single scenario assigned. For each video,
We select all time-stamps that contains a synchronized IMU signal as well as aligned narrations. We sample 5 second
clips around each time-stamp. The dataset is split randomly such that we have 510,142 clips for training, and 68,865
clips for testing. During training we only use the video frames and their corresponding IMU signal. We use the test
split to measure zero-shot scenario classification performance, where each clip of IMU signal is assigned the
video-level scenario label as its ground-truth.

\subsection{Data Representations}
We use the standard RGB and RGBT representations for {\bf images and videos}. For videos, we use 2-frame clips, inspired from recent work on ViT-style video architectures~\cite{tong2022videomae,feichtenhofer2022masked}, where a video patch is $2\!\times\!16\!\times\!16$ ($T\!\times\!H\!\times\!W$). We inflate the visual encoder's weights to work with spatiotemporal patches and and at inference time we aggregate features over multiple 2-frame clips. Hence, we can use models trained on image-text data directly on videos.

We used a single-channel image for the
{\bf thermal data} since it is the natural form in which current infrared thermal sensors return data~\cite{llvip}.
For
{\bf single-view depth}, we experimented with different encodings -- absolute depth~\cite{silberman2012indoor}
as
returned by sensors like the Kinect, inverse depth~\cite{ranftl2020towards},
disparity~\cite{ranftl2020towards}, and %
HHA~\cite{gupta2013perceptual,gupta2014learning}. Overall, we
found that disparity representation (which is a single-channel image) worked the best. For {\bf audio} we use the raw waveform processed into mel-spectrograms~\cite{gong21b_interspeech}, as described in the main text.
For {\bf IMU} we use a $6\!\times\!T$ tensor to represent the sequence of IMU sensor readings over time.

\section{Evaluation details}
\label{appendix:eval_details}

We now describe the evaluation setups used in this work.

\subsection{Inference implementation details}

{\bf \noindent Audio/Video:} For both these temporal modalities (whether operated upon together during pre-training or
separately during inference), we sample fixed length clips to operate on. During training, we randomly sample a clip,
typically 2s in length. At inference time, we uniformly sample multiple clips to cover the full length of the input
sample. For instance, for 5s \escShort videos, we would sample $\lceil \frac{5}{2} \rceil = 3$ clips. For video clips,
we sample a fixed number of frames from each clip. For audio, we process each raw audio waveform by sampling it at
16KHz followed by extracting a log mel spectrogram with 128 frequency bins using a 25ms Hamming window with hop length
of 10ms. Hence, for a $t$ second audio we get a $128\times\!100t$ dimensional input.

{\bf \noindent IMU:} For IMU, we sample fixed length clips of 5 seconds, centered around time-stamps that are aligned
with narrations. For each clip, we get a $6 \times 2000$ dimensional input and we measure the zero-shot performance for
scenario classification using each clip as an independent testing sample.

\subsection{Few-shot evaluation details}
\label{appendix:low_shot_eval}

\noindent For the few-shot results in Figures~\ref{fig:low_shot} using the \escShort and \sunrgbdShort datasets, we
sampled $k$ training samples per class, where $k \in \{1, 2, 4, 8\}$. We fix the $k$ samples such that our model and
the baselines use exactly the same samples during training. For all few-shot evaluations, including the baselines, we
freeze the encoder parameters and only train a linear classifier.

{\bf \noindent Audio:} For audio few-shot training with \escShort, our model and the baselines are trained using AdamW
with a learning rate of $1.6\times10^{-3}$ and weight decay of $0.05$ for 50 epochs.

{\bf \noindent Depth:} For depth few-shot training with \sunrgbdShort, our model and the baselines are trained using
AdamW with a learning rate of $10^{-2}$ and no weight decay for 60 epochs.

\subsection{Zero-shot evaluation details}
\label{appendix:zero_shot_eval}

{\bf \noindent Query Templates.}
For all evaluations, we use the default set of templates from CLIP~\cite{radford2021learning}.\footnote{\url{https://github.com/openai/CLIP/blob/main/notebooks/Prompt_Engineering_for_ImageNet.ipynb}} Note that we use the same templates for non visual modalities like audio and depth as well since we only use semantic/textual supervision associated with images.

\subsection{Qualitative evaluation details}
\label{appendix:qual_eval}

{\bf \noindent Cross-modal nearest neighbors.}
We perform the retrieval on the embedding feature after temperature scaling. The nearest neighbors are computed using cosine distance.
In~\cref{fig:teaser}, we show retrievals for audio from \escShort, image retrievals from \imnetShort and \cocoShort, depth from \sunrgbdDepthShort, and text from \audiocapsShort.

{\bf \noindent Embedding arithmetic.}
For arithmetic, we again use the embedding features after temperature scaling. We $\ell_2$ normalize the features and sum the embeddings
after scaling them by $0.5$. We use the combined feature to perform nearest neighbor retrieval using cosine distance, as described above. In~\cref{fig:teaser}, we show combination of images and audio from \imnetShort and \escShort, and show retrievals from \imnetShort.

{\bf \noindent Audio$\rightarrow$Image Generation.} For generating images form audio clips, we rely on an in-house reproduced implementation of DALLE-2~\cite{ramesh2022hierarchical}. In DALLE-2, to produce images from text prompts, the image generation model relies on text embeddings produced by the pre-trained CLIP-L/14 text encoder. Since \OURS naturally aligns CLIP's-embedding space to that of other modalities proposed in the paper, we can upgrade the DALLE-2 model to generate images by prompting it with these new unseen modalities. We achieve zero-shot audio to image generation with DALLE-2 by simply using the temperature-scaled audio embeddings generated by \OURS's audio encoder as a proxy for the CLIP's text embeddings in the DALLE-2's image generation model.

{\bf \noindent Detecting objects using audio.} We extract all audio descriptors from the validation set of \escShort using an \OURS ViT-B/32 encoder, yielding 400 descriptors in total. We use an off-the-shelf
CLIP-based Detic~\cite{zhou2022detecting} model and use the audio descriptors as the classifier for Detic in place of CLIP text-based `class' embeddings.
We use a score threshold of 0.9 for the qualitative results in Figure~\ref{fig:audio_detic}.

\section{Pretraining details}
\label{appendix:pretraining_details}

\subsection{Best setup}

In Table~\ref{tab:pretrain_settings} we detail the hyperparameters used to pre-train each of the models reported
in Table~\ref{sec:experiments}. Our experiments were done on 32GB V100 or 40GB A100 GPUs.

\begin{table}[!htb]
    \begin{center}
        \centering
        \resizebox{\linewidth}{!}{
        \begin{tabular}{l|cccc}
            Config & \color{AudioDark} \audiosetShort & \color{DepthDark} \sunrgbdShort & \color{ThermalDark} \llvipShort & \color{IMUDark}  \egoShort \\
            \midrule
            Vision encoder &  \multicolumn{4}{c}{ViT-Huge} \\
            embedding dim. & 768 & 384 & 768 & 512 \\
            number of heads & 12 & 8 & 12 & 8 \\
            number of layers & 12 & 12 & 12 & 6 \\
            Optimizer & \multicolumn{4}{c}{AdamW} \\
            Optimizer Momentum & \multicolumn{4}{c}{$\beta_1=0.9,
            \beta_2=0.95$} \\
            Peak learning rate & 1.6e-3 & 1.6e-3 & 5e-4 & 5e-4 \\
            Weight decay & 0.2 & 0.2 & 0.05 & 0.5 \\
            Batch size & 2048 & 512 & 512 & 512 \\
            Gradient clipping & 1.0 & 1.0 & 5.0 & 1.0 \\
            Warmup epochs & \multicolumn{4}{c}{2} \\
            Sample replication & 1.25 & 50 & 25 & 1.0 \\
            Total epochs & 64 & 64 & 64 & 8 \\
            Stoch. Depth~\cite{huang2016deep} & 0.1 & 0.0 & 0.0 & 0.7 \\
            Temperature & 0.05 & 0.2 & 0.1  & 0.2 \\
            Augmentations: \\
            \quad {\tt RandomResizedCrop} \\
            \qquad {\tt size} & \_ & \multicolumn{2}{c}{224px} & \_ \\
            \qquad {\tt interpolation} & \_ & Bilinear & Bilinear & \_ \\
            \quad {\tt RandomHorizontalFlip} & \_ & $p=0.5$ & $p=0.5$ & \_ \\
            \quad {\tt RandomErase} & \_ & $p=0.25$ & $p=0.25$ & \_ \\
            \quad {\tt RandAugment} & \_ & 9/0.5 & 9/0.5 & \_ \\
            \quad {\tt Color Jitter} & \_ & 0.4 & 0.4 & \_ \\
            \quad {\tt Frequency masking} & 12 & \_ & \_ & \_ \\
        \end{tabular}}
    \end{center}
    \caption{Pretraining hyperparameters}
    \label{tab:pretrain_settings}
    \end{table}

\paragraph{Contrastive loss batch size \vs modalities.} While contrastive losses do require larger
batch size, this requirement didn't increase with the number of modalities. As noted in~\cref{appendix:eval_details}, our
experiments (\cref{tab:emergent_zero_shot}) sample a mini-batch of one pair of modalities at a time: batch size of 2K for (video, audio), and 512 for (image, depth), (image, thermal), and
(video, IMU).
These batch sizes are smaller than the $>$32K batch sizes used in prior work~\cite{radford2021learning,chen2020simple}.

\paragraph{Combining modalities.} In~\cref{tab:text_retrieval_vtt}, we show results with combining the audio and video modalities. We combine them by extracting embeddings from both modalities per sample and computing a linear combinations of those embeddings. We used a weight of 0.95 for video and 0.05 for audio for this combination, which was found to perform the best.

\subsection{Ablation setup}
The following setup was used for our evaluations in~\cref{sec:ablation}. Different from the best setup, all ablation
experiments uses ViT-Base both for the vision and the modality-specific encoders. The models are trained for 16 epochs,
unless mentioned otherwise.

For~\cref{tab:ablate_proj_head}, the differences between the linear and MLP
heads are detailed below:
The MLP head did not improve performance in our experiments.
\begin{table}[h!]
  \resizebox{1.0\linewidth}{!}{
  \begin{tabular}{c | l}
    \toprule
    Linear & \texttt{Linear(in\_dim, out\_dim)} \\
    MLP & \texttt{Linear(in\_dim, in\_dim), GELU, Linear(in\_dim, out\_dim)} \\ %
    \bottomrule
  \end{tabular}
  }
\end{table}

\section{Additional Results}

{\noindent \bf Qualitative results.} We show additional results (along with audio) in the accompanying video.

\begin{figure}
    \centering
    \resizebox{\linewidth}{!}{
    \begin{tabular}{l}
      \textbf{Text query:} \textit{"Cooking a meal"} \\
      \includegraphics[width=0.65\textwidth]{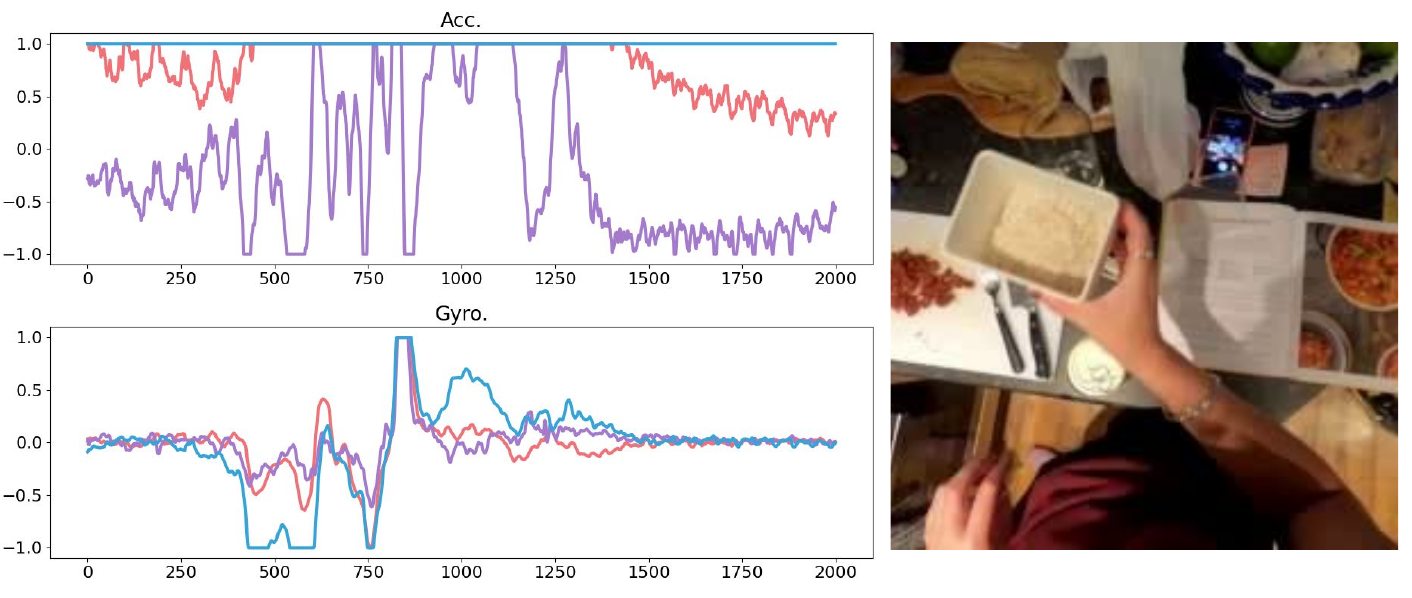} \\
      \textbf{Text query:} \textit{"A person doing gardening work outdoors"} \\
      \includegraphics[width=0.65\textwidth]{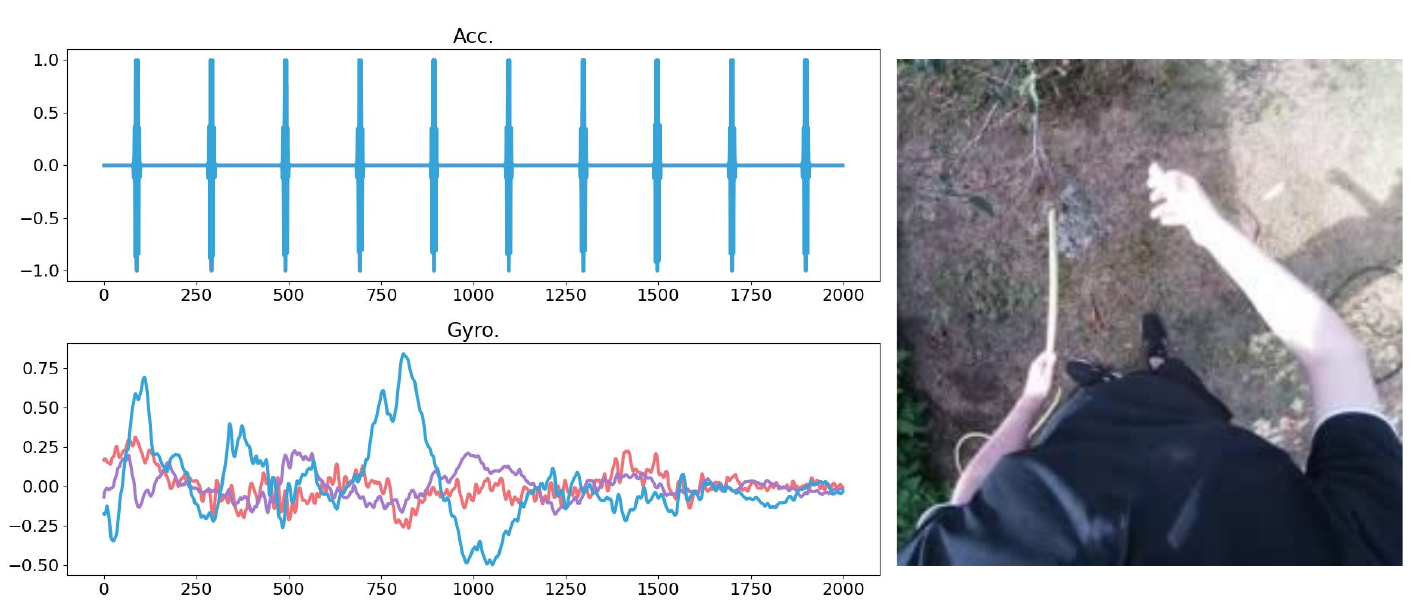} \\
    \end{tabular}
    }
    \caption{{\bf IMU retrievals.} Given a text query, we show some IMU retrievals and corresponding video frames.}
    \label{fig:appdx:imu}
  \end{figure}

{\bf \noindent Practical applications of disparate modalities.} In general, a shared embedding space enables a variety of
different cross-modal search and retrieval applications. \eg, since IMU sensors are ubiquitous (in phones,
AR/VR headsets, health trackers), \OURS can allow a user to search an IMU database using text queries (without
training with IMU-text pairs). IMU-based text search has applications in healthcare/activity search.
For instance, in~\cref{fig:appdx:imu} we show examples of IMU (and accompanying video) retrieval given textual search
query. The retrieved IMU sample, shown as 3-channel Accelerometer (Acc) and Gyroscope (Gyro) recording, matches the text query.

\section{Additional Ablations}

{\bf \noindent Design choices in losses.} Since the modality-specific encoders are trained to align
with a frozen image encoder, we tried using a $\ell_2$ regression objective. For ZS \sunrgbdShort top-1 accuracy, we observed
that regression led to good performance as the sole objective (25.17\%) or jointly with contrastive (29.04\%).
However, it did not improve over using only the contrastive objective (31.74\%).

\section{Ethical considerations}
\label{appendix:rai}
\OURS learns a joint embedding for multiple modalities.
Such an embedding is intended to associate semantically related concepts from different modalities.
However, such an embedding may also create unintentional associations.
Thus, joint embedding models, including \OURS must be studied carefully with a lens towards measuring such associations, and their implications.
\OURS leverages the image-text embeddings learned by a pretrained model on large web-based data which has biases as documented in different studies~\cite{radford2021learning}.
For learning joint embeddings for other modalities such as audio, thermal, depth, and IMU we leverage datasets mentioned in~\cref{appendix:data_details}.
These joint embeddings are thus limited to the concepts present in the datasets.
For example, the thermal datasets we used are limited to outdoor street scenes, while the depth datasets are limited to indoor scenes.

\end{document}